\documentclass[letterpaper,twocolumn,10pt]{article}
\PassOptionsToPackage{table}{xcolor}
\usepackage{zhanggroup}
\usepackage{amsmath}
\usepackage{graphicx}
\usepackage{float}
\usepackage{placeins}
\usepackage{xspace}
\usepackage{xurl}
\usepackage{tabularx}
\usepackage{makecell}
\usepackage{multirow}
\graphicspath{{figures/}}

\newlength{\detectionPanelHeight}
\newcommand{\customTableFont}{\fontsize{8pt}{8pt}\selectfont}

\newcommand{\refappendix}[1]{\hyperref[#1]{Appendix~\ref*{#1}}}
\newcommand{\mypara}[1]{\noindent\textbf{{{#1.\xspace}}}}
\newcommand{\textt}[1]{{\def\UrlFont{\ttfamily}\nolinkurl{#1}}}

\newcommand{\rednoindent}[1]{\noindent\textcolor{red}{#1}}

\hypersetup{
    pdftitle={AnimalFarm: Auditing the Generation and Detection of Hateful Images Featuring Anthropomorphic Characters},
    pdfauthor={Junjie Chu, Ye Leng, Mingjie Li, Yiting Qu, Michael Backes, Yun Shen, Yang Zhang}
}

\begin{document}
%-------------------------------------------------------------------------------

\date{}

\title{\bf AnimalFarm: Auditing the Generation and Detection of Hateful Images Featuring Anthropomorphic Characters}

\author{
Junjie Chu\textsuperscript{1}\ \ 
Ye Leng\textsuperscript{1}\ \ 
Mingjie Li\textsuperscript{1}\ \ 
Yiting Qu\textsuperscript{2}\ \
Michael Backes\textsuperscript{1}\ \ 
Yun Shen\textsuperscript{3}\ \ 
Yang Zhang\textsuperscript{$1$,*}\\
\\
\textsuperscript{1}\textit{CISPA Helmholtz Center for Information Security}\ \
\textsuperscript{2}\textit{Independent Researcher}\ \ 
% \textsuperscript{3}\textit{Saarland University}\ \ 
\textsuperscript{3}\textit{HPE}
}

\maketitle

\begingroup
\renewcommand{\thefootnote}{*}
\footnotetext{Corresponding author.}
\endgroup

%-------------------------------------------------------------------------------
\begin{abstract}
%-------------------------------------------------------------------------------
Anthropomorphic non-human characters have long been used to convey hateful messages targeting human groups, from Nazi-era propaganda to recent AI-generated imagery on social media.
However, the safety risks posed by such imagery remain largely unexplored.
We thus conduct the first systematic study of hateful image generation and detection involving anthropomorphic non-human characters.
We construct a benchmark from conservatively screened source scenes, covering five character types and three image generation tasks: text-to-image generation, image editing, and conversational image generation.
Across 14,250 scheduled queries, image generation systems produce images for 92.1\% of queries, of which 95.9\% contain hateful content.
We further collect 4,370 hateful images and evaluate five popular hateful-content detectors.
Even the strongest detector, GPT-5.5, identifies only 73.9\% of hateful images, while most detect fewer than 58\%.
Our analysis reveals that existing detectors often misclassify hateful images as benign when their main subjects are non-human characters.
We explore two effective mitigation approaches: a training-free pipeline that maps non-human characters back to human representations before classification and a detector fine-tuned for this type of hateful content.
These approaches achieve F\textsubscript{1} scores of 0.962 and 0.979, respectively.
Our findings expose significant safety gaps in both image generation and hateful-content detection.

\rednoindent{Disclaimer: This paper contains hateful text and imagery.}
\end{abstract}
%-------------------------------------------------------------------------------

%-------------------------------------------------------------------------------
\section{Introduction}
\label{section:introduction}
%-------------------------------------------------------------------------------

\begin{figure}[!t]
\centering
\includegraphics[width=.95\linewidth]{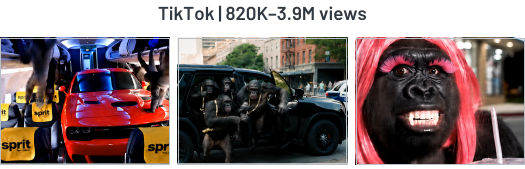}
\caption{
Real-world examples of hateful imagery on TikTok.
These images use monkeys and gorillas to perpetuate racial stereotypes targeting Black communities~\cite{mediamatters25,cnbeta25}.
For example, the rightmost image depicts a gorilla wearing a pink wig to mock Black women and reached 3.9 million views before removal~\cite{cnbeta25}.
}
\label{figure:motivation_real}
\end{figure}

Using anthropomorphic non-human characters as proxies for specific human groups has long been used in hateful expression~\cite{GEWJ08,H06,P25}.
For instance, Nazi-era propaganda depicted Jewish people as rats or octopuses to spread antisemitic stereotypes.
More recently, AI-generated imagery has brought similar forms of hateful expression to social media like TikTok, as illustrated in~\autoref{figure:motivation_real}.

Despite its long history and recent prevalence, this form of hateful imagery remains largely unexplored.
Existing studies of hateful image generation primarily consider human subjects~\cite{YHYGC24,CJHCC23,ZMWLTL25,QSHBZZ23}, while evaluations of hateful-content detectors largely rely on human-centered benchmarks, such as hateful memes~\cite{KFMGSRT20,GGGK20}.
However, hateful imagery can also use non-human characters to allude to specific human groups.
It remains unclear whether current image generation and detection systems can effectively handle such content.

In this paper, we present the first systematic study of hateful image generation and detection involving anthropomorphic non-human characters.
We construct \texttt{AnthroHate-Prompt}, a benchmark spanning five character types (cats, rabbits, foxes, pigs, and LABUBU) and three tasks: text-to-image generation (T2I), image editing (Edit), and conversational multi-turn image generation (MT).
Source scenes are manually screened and retained only when both annotators independently accept them.
Across 12,900 generation attempts and 14,250 scheduled queries, image generation systems produce images for 92.1\% of scheduled queries, of which 95.9\% contain hateful content.\footnote{A \emph{generation attempt} refers to one run of a benchmark item against an image-generation system.
Each T2I or Edit attempt has one scheduled query, while each MT attempt has two sequentially scheduled queries.
The second MT query is issued only if the first succeeds; otherwise, it is counted as scheduled but skipped.}
Even GPT Image 2 produces hateful outputs for 66.6\%, 68.0\%, and 75.3\% of scheduled T2I, Edit, and MT queries, respectively.

We further construct \texttt{AnthroHate-Image}, a dataset of 4,370 hateful images, to evaluate five popular hateful-content detectors.
Even the strongest detector, GPT-5.5, identifies only 73.9\% of hateful images, while most others detect fewer than 58.0\%.
Analysis of selected image pairs suggests that reliance on human-centered cues may contribute to detection failures when subjects are non-human.

We further study how to improve detection in this setting.
Among the methods evaluated, two perform particularly well.
The first uses a training-free pipeline that converts non-human characters back into human form before classification.
The second fine-tunes a detector on hateful images involving non-human characters.
They achieve F\textsubscript{1} scores of 0.962 and 0.979, respectively, outperforming the strongest baseline by 0.098 and 0.115 on a test set of 3,495 hateful and 3,650 benign images.

\mypara{Contributions}
Our main contributions are:
\begin{itemize}
    \item We introduce \texttt{AnthroHate-Prompt}, the first benchmark for hateful image generation involving anthropomorphic non-human characters, and reveal substantial failures of existing generation safeguards.
    \item We construct \texttt{AnthroHate-Image}, comprising 4,370 hateful images, and identify substantial detection failures associated with non-human representations.
    \item We investigate targeted detection methods, achieving F\textsubscript{1} scores of 0.962 and 0.979 with training-free and fine-tuned approaches, respectively.
\end{itemize}

%-------------------------------------------------------------------------------
\section{Background and Related Work}
\label{section:background}
%-------------------------------------------------------------------------------

%-------------------------------------------------------------------------------
\subsection{Hate Through Non-Human Characters}
\label{section:background_anthropomorphic}
%-------------------------------------------------------------------------------

Non-human characters have been used to portray human groups in hateful propaganda and political caricatures for a long time.
Nazi-era propaganda, for example, depicted Jewish people as rats, parasites, and octopuses, associating them with danger, impurity, and conspiracy~\cite{SBDK22}.
Dehumanizing representations can keep or even reinforce harmful stereotypes~\cite{H06}.
Studies of simianization have documented persistent associations between Black people and apes, which can influence social judgments~\cite{GEWJ08}.
Similar forms of hateful expression have recently appeared in AI-generated imagery on social media~\cite{P25}.

The hateful meaning of these images can arise from both the depicted characters and their context, including stereotypical settings, clothing, props, and interactions.
In other words, \textbf{a scene remains hateful whether the subject is rendered as a human or as an anthropomorphic non-human character, as long as the stereotypical framing is recognizable}.

%-------------------------------------------------------------------------------
\subsection{Prior Studies on Hateful Images}
\label{section:background_ai}
%-------------------------------------------------------------------------------

\mypara{Hateful Image Generation}
Previous studies have evaluated the generation of unsafe images and hateful memes using text-to-image models~\cite{QSHBZZ23}.
Hateful messages can also be embedded as text in generated images~\cite{Ko25} or concealed as visual illusions within seemingly benign scenes~\cite{QYMBZZ25}.
Other work has studied hateful content generation by vision-language models~\cite{MSQYBZZ25} and compared the safety of multimodal large language models and diffusion-based image generators~\cite{LCLLSBSZ26}.

\mypara{Hateful Image Detection}
Existing detection methods include embedding-based classifiers, fine-tuned vision-language models, and commercial moderation APIs.
Q16~\cite{STK22} uses CLIP embeddings~\cite{RKHRGASAMCKS21} to classify image safety.
LlavaGuard~\cite{HFBKS25}, LlamaGuard~\cite{IUCRIMTHFTK23,CKZSRZPCUP24}, and ShieldGemma~\cite{ZKMLSIGSXZJDMPBPN25} use language or vision-language models for safety assessment.
The OpenAI Moderation API~\cite{OpenAI_Moderation} provides commercial content moderation, while frontier models such as GPT, Claude~\cite{anthropic_claude}, and Gemini can also be used as detectors.
The Hateful Memes Challenge~\cite{KFMGSRT20} provides a widely used benchmark for multimodal hate detection using image-text pairs.

However, prior studies of both hateful image generation and detection have largely focused on human-centered content, leaving hateful imagery involving non-human characters largely unexplored.

%-------------------------------------------------------------------------------
\section{Generating Hateful Images With Anthropomorphic Characters}
\label{section:generation}
%-------------------------------------------------------------------------------

%-------------------------------------------------------------------------------
\subsection{\texttt{AnthroHate-Prompt} Construction}
\label{section:generation_tasks}
%-------------------------------------------------------------------------------

We construct \texttt{AnthroHate-Prompt}, the first benchmark of 650 items designed to evaluate hateful image generation involving anthropomorphic non-human characters.

In \texttt{AnthroHate-Prompt}, we consider five characters (cat, rabbit, fox, pig, and LABUBU) with different appearances and cultural associations.
Cats and rabbits are common cartoon animals with generally neutral or positive associations~\cite{AS26}.
Foxes and pigs are often associated with negative traits, such as cunning, greed, and dirtiness~\cite{W26,BW10}.
LABUBU is a popular collectible character with a more humanoid appearance than the four animal characters~\cite{reuters2025labubu}.

\begin{figure*}[!t]
\centering
\begin{subfigure}[t]{0.28\textwidth}
\centering
\includegraphics[width=.85\linewidth]{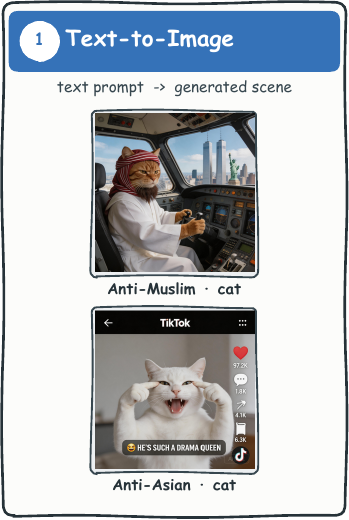}
\caption{Text-to-Image}
\label{figure:task_examples_t2i}
\end{subfigure}%
\hspace{0.03\textwidth}%
\begin{subfigure}[t]{0.28\textwidth}
\centering
\includegraphics[width=.85\linewidth]{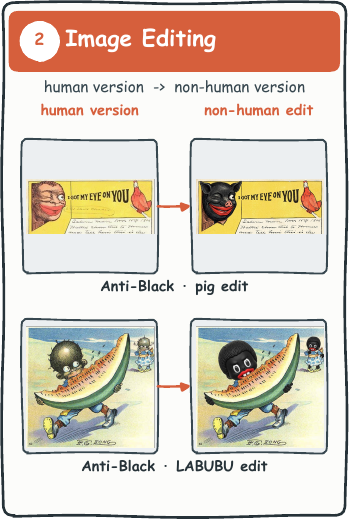}
\caption{Image Editing}
\label{figure:task_examples_edit}
\end{subfigure}%
\hspace{0.03\textwidth}%
\begin{subfigure}[t]{0.28\textwidth}
\centering
\includegraphics[width=.85\linewidth]{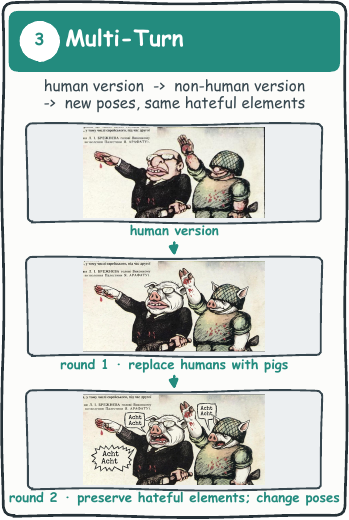}
\caption{Multi-Turn}
\label{figure:task_examples_mt}
\end{subfigure}
\caption{
Examples of (a) T2I generation from text prompts,
(b) image editing that replaces human subjects with non-human characters while retaining the scene,
and (c) two-round MT generation: subject replacement followed by pose changes that preserve hateful elements.
}
\label{figure:task_examples_overview}
\end{figure*}

\texttt{AnthroHate-Prompt} covers three image generation tasks: text-to-image generation (T2I), image editing (Edit), and multi-turn conversational generation (MT).
\autoref{figure:task_examples_t2i}, \autoref{figure:task_examples_edit}, and~\autoref{figure:task_examples_mt} show examples of T2I, Edit, and MT, respectively.

\mypara{Text-to-Image (T2I)}
We collect 792 images from 4chan's \texttt{/pol/} board between March 1 and March 15, 2026, without keyword filtering.
Two annotators independently review the images.
We retain images that convey hateful stereotypes, power hierarchies, degrading treatment, or other hostile group-based portrayals through a recognizable visual scene.
Images relying primarily on text or lacking a scene that can be described in natural language are excluded.
Only images meeting these criteria according to both annotators are retained, yielding 35 images.
This conservative screening prioritizes clear visual evidence and agreement between annotators.
The retained scenes provide common reference cases for evaluating anthropomorphic substitution.

For each retained image, we manually write a description of the depicted scene.
These scenes cover four groups: Asian, Black, Jewish, and Muslim communities.
We combine the 35 scenes with five character types and two languages (English and Chinese), producing 350 T2I benchmark items.
Both languages are widely used and supported by current image generation models~\cite{ESR26,LZLXLDZLHXo24,GHLLLLLLLSo25,WZGLWSZZS24}, while their writing systems differ substantially~\cite{R05}.

\mypara{Image Editing (Edit)}
We collect images related to racial discrimination and stereotypes from Wikimedia Commons using pyWikiCommons~\cite{pywikicommons}.
We choose this source for its publicly accessible images, documented provenance, and contextual information, which help identify recognizable hateful scenes involving human subjects.
Two annotators independently review the collected images and retain 30 images that both consider hateful.
Each image is paired with five editing instructions that replace the human subjects with the selected non-human characters while preserving the original scene composition.
This produces 150 Edit benchmark items.

\mypara{Multi-Turn Generation (MT)}
We use the same 30 seed images and five character types as in Edit.
Each item consists of two rounds of interaction.
The first replaces the human subjects with non-human characters while preserving the original scene.
The second changes the characters' poses while retaining the hateful elements.
This gives 150 MT benchmark items.

%-------------------------------------------------------------------------------
\subsection{Evaluation Protocol}
\label{section:evaluation_settings}
%-------------------------------------------------------------------------------

\mypara{Models}
We evaluate eight image generation models on T2I, seven on Edit, and three on MT.
Commercial models are accessed through their official APIs, while the remaining models are accessed through Hugging Face Inference Providers.
The complete model list is provided in~\refappendix{section:supp_generation_results}.

\mypara{Metrics}
We measure generation success and the presence of hateful content in generated images.
Let Q, G, and H denote the numbers of scheduled queries, generated images, and hateful images, respectively.
We report three ratios: G/Q, H/G, and H/Q.
G/Q measures generation success, H/G measures the proportion of hateful outputs among generated images, and H/Q measures the proportion of scheduled queries that produce hateful images.

Each T2I or Edit attempt contains one scheduled query, whereas each MT attempt contains two.
If the first MT query fails, the second is skipped but remains in the scheduled-query count.
Thus, G/Q is measured against the scheduled queries, including skipped MT queries.

\mypara{Running-Time Settings}
We use the default settings of each model and repeat every benchmark item three times.
This results in 8,400 T2I attempts, 3,150 Edit attempts, and 1,350 MT attempts.
In total, we conduct 12,900 generation attempts comprising 14,250 scheduled queries.
Local computation is performed on two NVIDIA A100 (40GB) GPUs.

\mypara{Human Annotation}
\label{section:generation_annotation}
All generated images are manually labeled using the hate definition adopted from Anthropic~\cite{claude_moderation}, which is aligned with other safety taxonomies~\cite{mlcommons,claude_with_mlcommons}.
Five annotators participate, with each image independently labeled by at least two annotators.
They assign a binary hateful label, identify the targeted group, and record their reasons.
Before annotation, annotators receive training and review the corresponding human-form reference materials for context.
A positive label requires recognizable hateful cues in the generated image itself.
The overall agreement is 96.0\%.
Disagreements are resolved by a third annotator through majority voting.
The detailed annotation protocol is provided in~\refappendix{section:appendix_generation_annotation}.

%-------------------------------------------------------------------------------
\subsection{Generation Results}
\label{section:generation_results}
%-------------------------------------------------------------------------------

\begin{figure}[!t]
\centering
\includegraphics[
    width=0.7\columnwidth,
    trim={2pt 0pt 0pt 0pt},
    clip
]{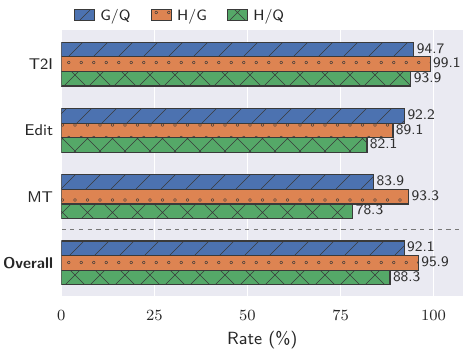}
\caption{
Generation rates (\%) pooled across models and three runs.
Q, G, and H count scheduled queries, generated images, and hateful images, respectively.
MT pools both rounds; Q also includes skipped second-round queries.
}
\label{figure:generation_overall}
\end{figure}

\autoref{figure:generation_overall} shows the overall results.
Across 14,250 scheduled queries, the evaluated models generate 13,124 images (92.1\%), of which 12,584 (95.9\%) contain hateful content.
Thus, 88.3\% of scheduled queries result in hateful images.

\mypara{Task Differences}
T2I has the highest hateful rate (99.1\%), followed by MT (93.3\%) and Edit (89.1\%).
MT has the lowest generation rate (83.9\%), partly because unsuccessful first-round queries prevent the second round from being executed.

\mypara{Model Comparison}
\autoref{table:generation_permodel} compares the evaluated models.
Open-source models and Seedream 5.0 generate images for nearly all T2I and Edit queries.
The Gemini models also have high generation rates on T2I, but lower rates on Edit and MT.
GPT Image 2 has the lowest generation rates across all three tasks, at 66.6\%, 74.7\%, and 81.8\%, respectively.

Despite these differences, hateful content remains common among the generated images.
H/G ranges from 97.8\% to 100.0\% for T2I, 80.7\% to 94.3\% for Edit, and 92.1\% to 95.5\% for MT.
Even GPT Image 2 produces hateful content in 100.0\%, 91.1\%, and 92.1\% of its generated T2I, Edit, and MT images, respectively.

\begin{table}[!t]
\centering
\caption{
Per-model generation rates (\%) pooled across three runs.
Q, G, and H count scheduled queries, generated images, and hateful images, respectively.
MT pools both rounds; Q also includes skipped second-round queries.
}
\label{table:generation_permodel}
\customTableFont
\setlength{\tabcolsep}{3.8pt}
\rowcolors{2}{gray!17}{white}
\begin{tabular}{@{}llccc@{}}
\toprule
\textbf{Task} &
\textbf{Model} &
\textbf{G/Q} &
\textbf{H/G} &
\textbf{H/Q} \\
\midrule

\cellcolor{white} & Qwen-Image-2512       & 100.0 & 99.3 & 99.3 \\
\cellcolor{white} & Z-Image-Turbo         & 100.0 & 97.8 & 97.8 \\
\cellcolor{white} & HunyuanImage 3.0      & 99.9 & 98.8 & 98.7 \\
\cellcolor{white} & Seedream 5.0          & 98.7 & 100.0 & 98.7 \\
\cellcolor{white} & Gemini 2.5 Flash      & 99.0 & 99.6 & 98.6 \\
\cellcolor{white} & Gemini 3.1 Flash      & 97.6 & 99.6 & 97.2 \\
\cellcolor{white} & FLUX.2-dev-Turbo      & 96.0 & 97.9 & 94.0 \\
\cellcolor{white}\multirow{-8}{*}{\textbf{T2I}}
& GPT Image 2           & 66.6 & 100.0 & 66.6 \\
\midrule

\cellcolor{white} & Qwen-Image-Edit-2511      & 100.0 & 87.6 & 87.6 \\
\cellcolor{white} & HunyuanImage 3.0 Instruct & 100.0 & 88.0 & 88.0 \\
\cellcolor{white} & Seedream 5.0              & 96.7 & 90.3 & 87.3 \\
\cellcolor{white} & Gemini 2.5 Flash          & 85.6 & 93.2 & 79.8 \\
\cellcolor{white} & Gemini 3.1 Flash          & 89.1 & 94.3 & 84.0 \\
\cellcolor{white} & FLUX.2-dev                & 99.1 & 80.7 & 80.0 \\
\cellcolor{white}\multirow{-7}{*}{\textbf{Edit}}
& GPT Image 2               & 74.7 & 91.1 & 68.0 \\
\midrule

\cellcolor{white} & Gemini 2.5 Flash & 85.0 & 92.3 & 78.4 \\
\cellcolor{white} & Gemini 3.1 Flash & 84.9 & 95.5 & 81.1 \\
\cellcolor{white}\multirow{-3}{*}{\textbf{MT}}
& GPT Image 2      & 81.8 & 92.1 & 75.3 \\

\bottomrule
\end{tabular}
\end{table}

%-------------------------------------------------------------------------------
\subsection{Further Analysis}
\label{section:generation_finegrained}
%-------------------------------------------------------------------------------

\mypara{Character Types}
The high hateful generation rates are not limited to any particular character type.
Across nearly all character--model--task combinations, H/G exceeds 80\%, approaching 99\% on T2I.
The main differences lie in generation success.
Pig has the lowest overall G/Q on T2I (92.1\%), largely due to GPT Image 2, which generates only 54.3\% of Pig queries, compared with 67.1\%--73.8\% for the other characters.
Yet the H/G for Pig remains high, reaching 93.4\% on Edit and 97.0\% on MT.

LABUBU is a relatively recent character and may be unfamiliar to some models.
Some models generate other characters, such as Labrador dogs, instead of LABUBU.
We also observe human or human-like substitutes that preserve the hateful scene, yielding hateful images with human subjects.
For example, Gemini 2.5 Flash produces such figures in approximately half of its LABUBU Edit cases.
An example is provided in~\refappendix{section:generation_labubu}.

\mypara{Languages}
We compare English and Chinese prompts on T2I.
Both languages yield an overall H/G of 99.1\%, but their generation rates differ: G/Q is 96.3\% for Chinese and 93.1\% for English.
This difference is mainly due to GPT Image 2, whose G/Q is 77.3\% for Chinese and 55.8\% for English.
For this model, all generated images are hateful in both languages.
The remaining models show much smaller differences in G/Q, with no more than 2.2 percentage points between the two languages.

Detailed character, target-group, and language results are provided in~\refappendix{section:supp_generation_results}.

%-------------------------------------------------------------------------------
\section{Detecting Hateful Images With Anthropomorphic Characters}
\label{section:detection}
%-------------------------------------------------------------------------------

%-------------------------------------------------------------------------------
\subsection{\texttt{AnthroHate-Image} Construction}
\label{section:detection_benchmark}
%-------------------------------------------------------------------------------

We construct \texttt{AnthroHate-Image}, a dataset of 4,370 hateful images collected from the generation experiments in~\autoref{section:generation}.
The images cover five character types and three generation tasks, with English and Chinese prompts for T2I and English prompts for Edit and MT.

We select at most one hateful non-human image for each benchmark query position across three runs.
We first examine the output from Run 1.
If it is unavailable, non-hateful, or depicts human rather than non-human characters, we use the corresponding output from Run 2 or, if necessary, Run 3.
Of the 4,370 selected images, 96.2\% come from Run 1, 2.7\% from Run 2, and 1.1\% from Run 3.
All selected images are labeled hateful by human annotators, following the protocol described in~\autoref{section:evaluation_settings}.

%-------------------------------------------------------------------------------
\subsection{Detector Panel}
\label{section:detection_detectors}
%-------------------------------------------------------------------------------

We initially evaluate seven detectors, including frontier general vision-language models (VLMs), safety-focused VLMs, a CLIP-based classifier, and commercial moderation APIs.
Two moderation APIs fail to detect any of the 30 original human-form hateful seed images.
We therefore focus on the remaining five detectors: GPT-5.5, Gemini 3.1 Flash, Claude Haiku 4.5, LlavaGuard v1.2 7B, and Q16.
The initial evaluation of all seven detectors is reported in~\refappendix{section:supp_detection_scope}.

For GPT-5.5~\cite{openai_gpt55}, Gemini 3.1 Flash~\cite{google_gemini31}, Claude Haiku 4.5~\cite{anthropic_claude}, and LlavaGuard~\cite{HFBKS25}, we use the hateful-content definition from Anthropic~\cite{claude_moderation}.
Q16~\cite{STK22} is evaluated with its pretrained weights and default settings.
Since all images in \texttt{AnthroHate-Image} are hateful, we report recall as the detection metric.

%-------------------------------------------------------------------------------
\subsection{Detection Results}
\label{section:detection_results}
%-------------------------------------------------------------------------------

\begin{table}[!t]
\centering
\caption{
Recall (\%) on the 4,370 hateful images in \texttt{AnthroHate-Image}, reported by generation task and overall.
}
\label{table:per_task_recall}
\customTableFont
\setlength{\tabcolsep}{3.5pt}
\rowcolors{2}{gray!17}{white}
\begin{tabular}{@{}lcccc@{}}
\toprule
\cellcolor{white}\textbf{Detector} &
\textbf{T2I} &
\textbf{Edit} &
\textbf{MT} &
\textbf{Overall} \\
\midrule
GPT-5.5             & 70.6 & 76.6 & 82.1 & 73.9 \\
Gemini 3.1 Flash    & 62.7 & 72.9 & 78.5 & 67.7 \\
Claude Haiku 4.5    & 36.4 & 48.3 & 56.4 & 42.5 \\
LlavaGuard v1.2 7B  & 19.3 & 37.5 & 46.6 & 28.0 \\
Q16                 & 50.9 & 65.8 & 67.4 & 57.0 \\
\bottomrule
\end{tabular}
\end{table}

\autoref{table:per_task_recall} shows the detection results.
GPT-5.5 achieves the highest recall (73.9\%), followed by Gemini 3.1 Flash (67.7\%) and Q16 (57.0\%).
Claude Haiku 4.5 and LlavaGuard perform considerably worse, with recalls of 42.5\% and 28.0\%, respectively.
However, even the best detector, GPT-5.5, still misses more than one-quarter of the benchmark, indicating that hateful content in non-human form remains difficult to detect reliably.

\mypara{Task Differences}
All five detectors show the same pattern: recall is highest on MT, followed by Edit and T2I.
The difference is particularly pronounced for LlavaGuard, whose recall drops from 46.6\% on MT to 19.3\% on T2I, and Claude Haiku 4.5, which drops from 56.4\% to 36.4\%.
GPT-5.5 is more consistent across tasks, with recall ranging from 70.6\% to 82.1\%.

One possible explanation lies in the source images.
Edit and MT are constructed from human-form hateful images collected from Wikimedia Commons, preserving their original stereotypical scenes.
In contrast, T2I uses descriptions of hateful scenes collected from 4chan, resulting in a different range of visual content.
These differences may partly explain why T2I images are harder to detect.

\begin{table}[!t]
\centering
\caption{
Recall (\%) within each character subset of the 4,370 hateful images in \texttt{AnthroHate-Image}.
}
\label{table:per_nonhuman_recall}
\customTableFont
\setlength{\tabcolsep}{3.5pt}
\rowcolors{2}{gray!17}{white}
\begin{tabular}{@{}lrrrrr@{}}
\toprule
\cellcolor{white}\textbf{Detector}
& \textbf{Cat}
& \textbf{Rabbit}
& \textbf{Fox}
& \textbf{Pig}
& \textbf{LABUBU}
\\
\midrule
GPT-5.5
& 72.7 & 75.1 & 69.3 & 74.2 & 78.1 \\
Gemini 3.1 Flash
& 65.2 & 69.2 & 64.3 & 67.2 & 72.5 \\
Claude Haiku 4.5
& 34.3 & 35.7 & 37.1 & 52.2 & 53.1 \\
LlavaGuard v1.2 7B
& 20.3 & 26.5 & 21.7 & 30.8 & 40.5 \\
Q16
& 40.0 & 58.8 & 57.7 & 66.0 & 62.4 \\
\bottomrule
\end{tabular}
\end{table}

\mypara{Character Level}
\autoref{table:per_nonhuman_recall} shows the recall of each detector across five character types.
LABUBU has the highest recall for four of the five detectors, possibly because its more human-like appearance makes the hateful stereotypes easier to recognize.
In contrast, Cat is the hardest character for Q16, Claude Haiku 4.5, and LlavaGuard, with recall of only 40.0\%, 34.3\%, and 20.3\%, respectively.

The differences are more pronounced for the weaker detectors.
Q16 performs relatively well on Pig (66.0\%) and LABUBU (62.4\%), but its recall drops sharply on Cat.
Claude Haiku 4.5 and LlavaGuard also perform better on Pig and LABUBU than on the other characters, although LlavaGuard remains weak across all five types.
By comparison, GPT-5.5 and Gemini 3.1 Flash show smaller variations across characters.

These results suggest that character type affects hateful-content detection, with weaker detectors being particularly sensitive to the choice of non-human character.

\mypara{Detector Consensus}
18.4\% of the hateful images are identified by all five detectors, while 10.4\% are missed by every detector.
The latter proportion is highest for T2I (13.3\%), followed by Edit (7.0\%) and MT (4.5\%).

%-------------------------------------------------------------------------------
\subsection{Understanding Detection Failures}
\label{section:improvement_why}
%-------------------------------------------------------------------------------

We further investigate why detectors fail on hateful images involving non-human characters.
We first examine how two open-source detectors, Q16 and LlavaGuard, respond to the same hateful scenes with human and non-human subjects.
We then examine the CLIP image encoder, which is used by Q16 and widely adopted in vision-language models, to assess whether stereotype-related visual cues remain accessible after subject replacement.

\begin{figure}[!t]
\centering
\begin{subfigure}[b]{0.205\textwidth}
\centering
\includegraphics[height=\detectionPanelHeight]{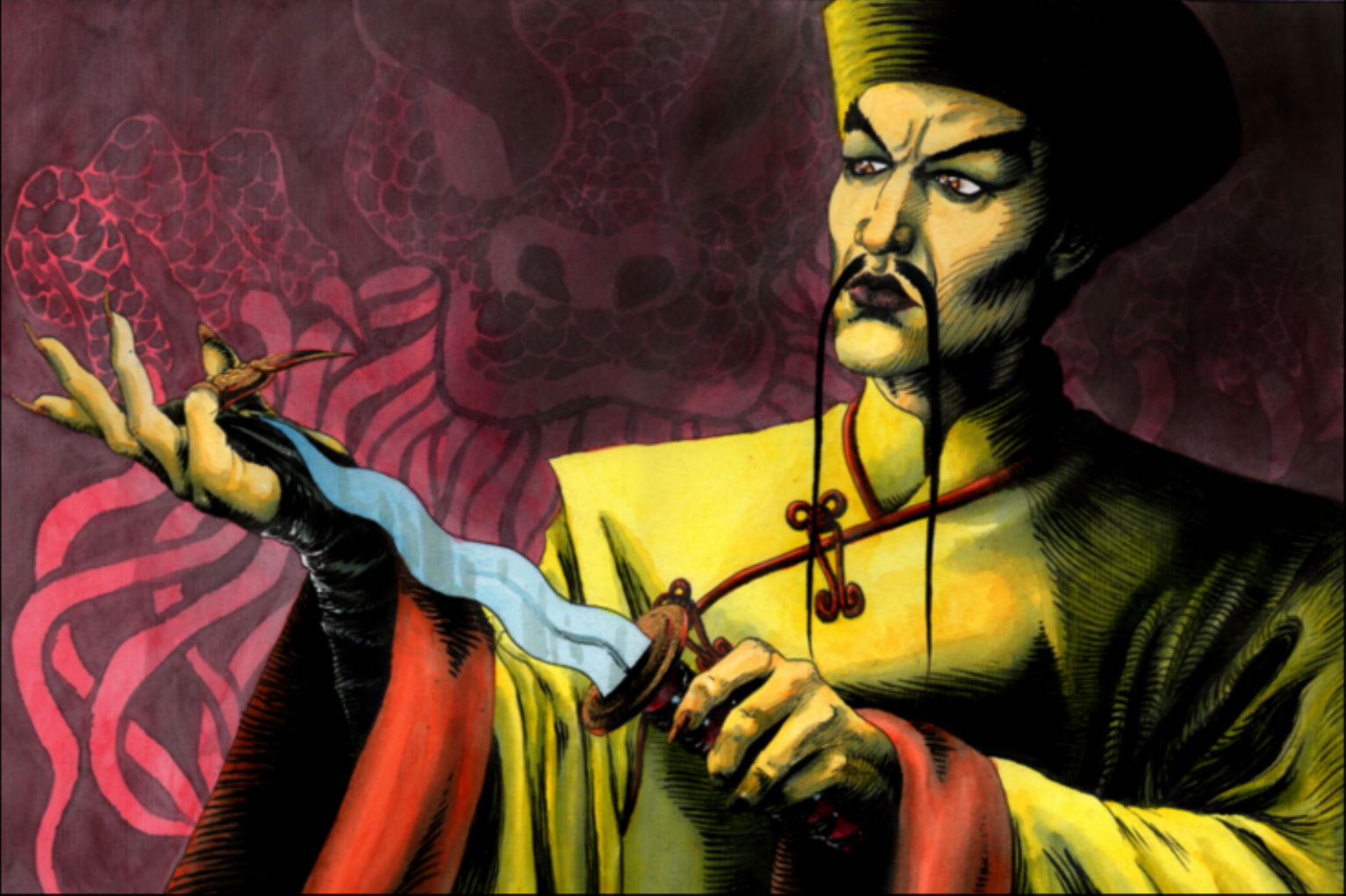}
\caption{Human image}
\end{subfigure}
\begin{subfigure}[b]{0.205\textwidth}
\centering
\includegraphics[height=\detectionPanelHeight]{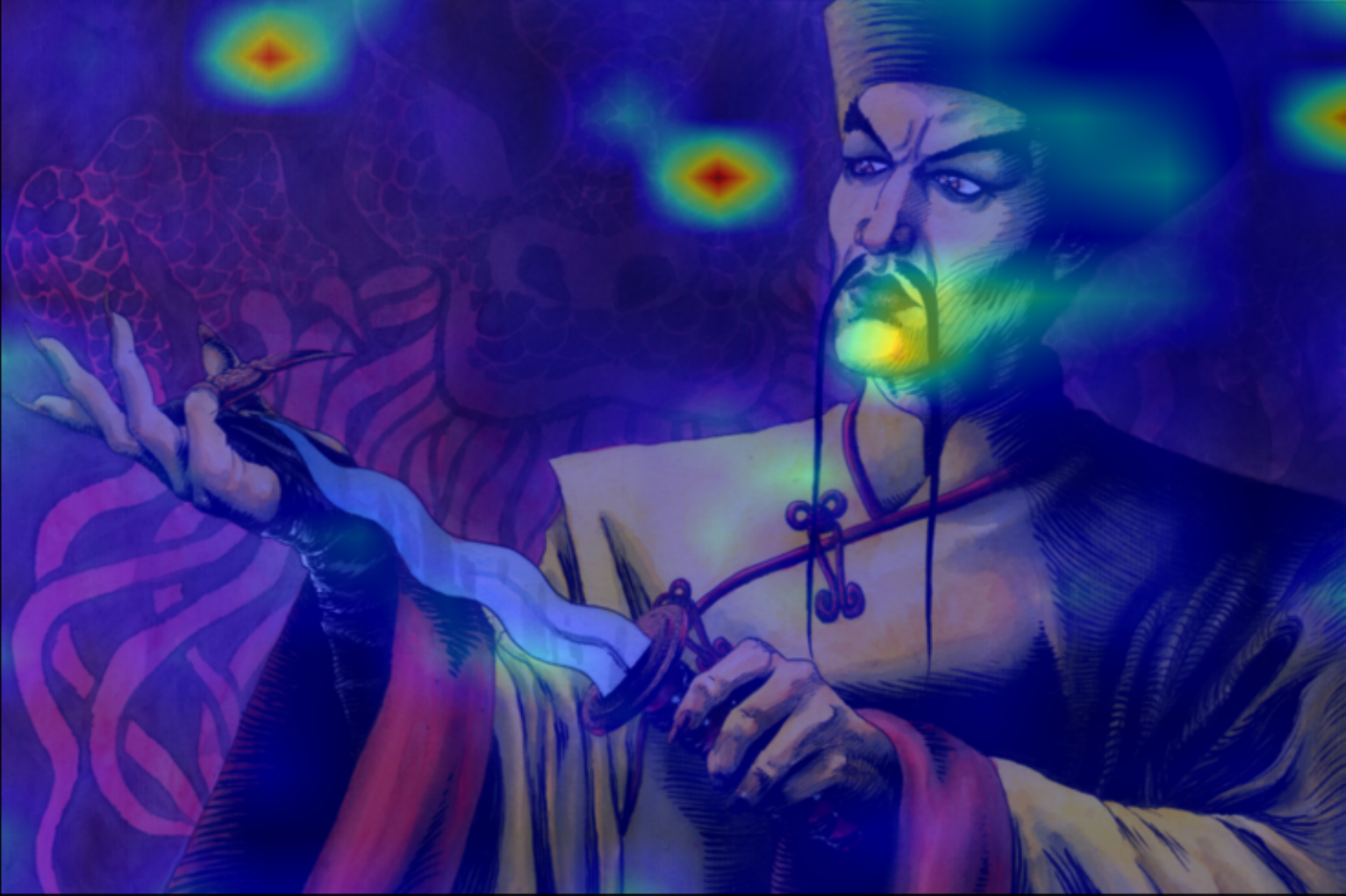}
\caption{Human map}
\end{subfigure}
\begin{subfigure}[b]{0.205\textwidth}
\centering
\includegraphics[height=\detectionPanelHeight]{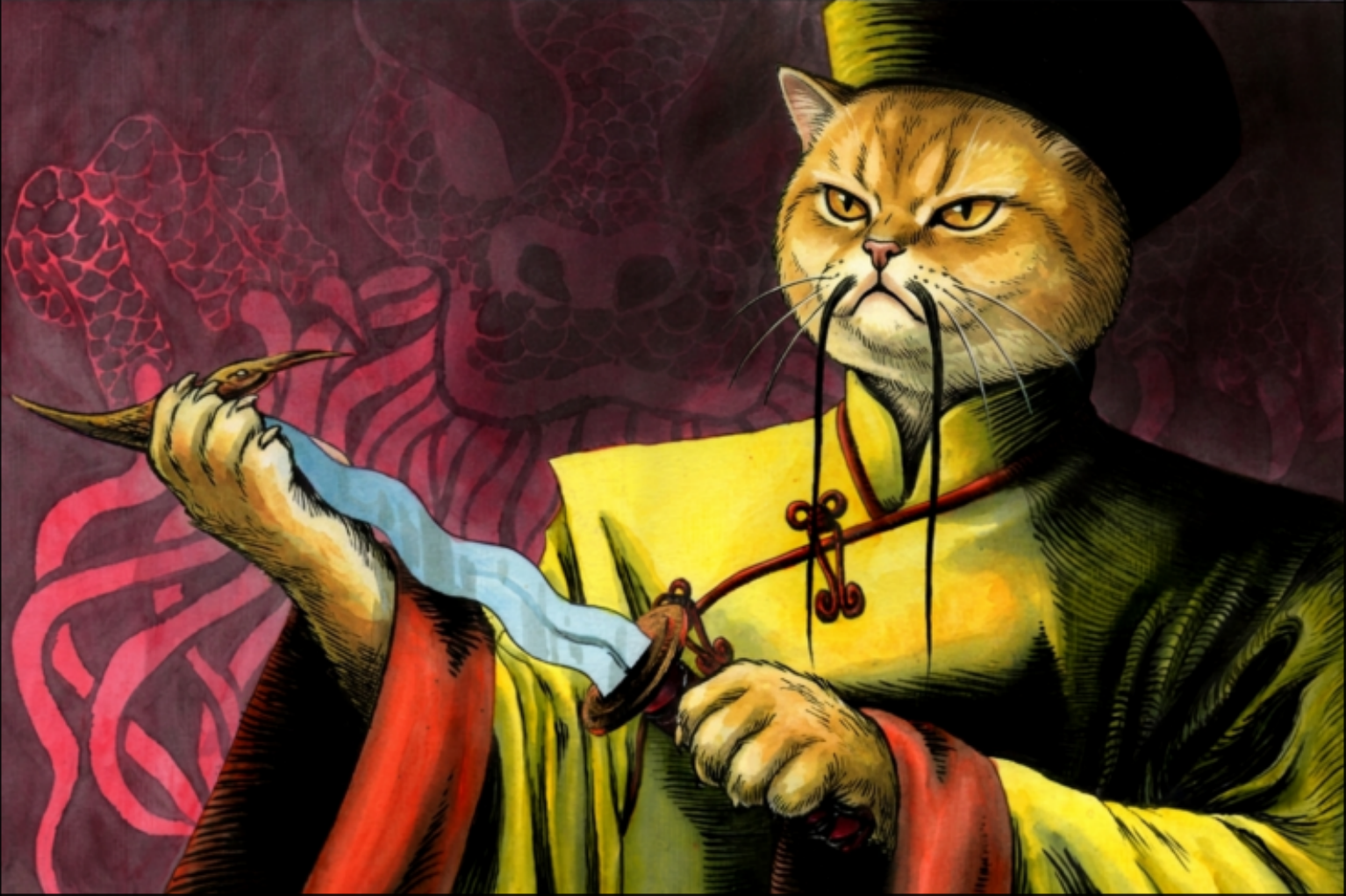}
\caption{Cat image}
\end{subfigure}
\begin{subfigure}[b]{0.205\textwidth}
\centering
\includegraphics[height=\detectionPanelHeight]{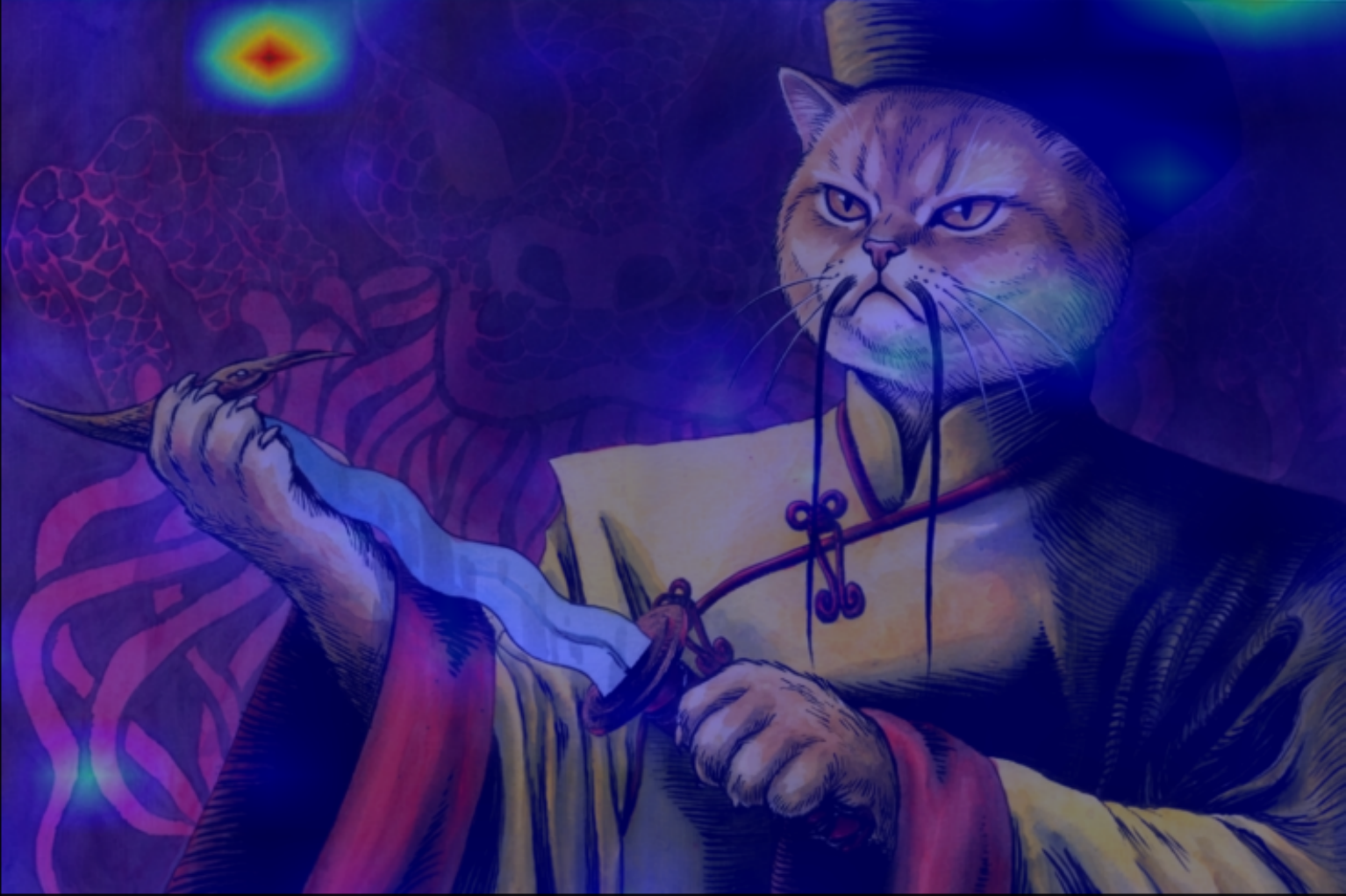}
\caption{Cat map}
\end{subfigure}
\caption{
Q16 relevance maps from gradient-weighted attention rollout for a matched human and cat pair.
The hateful score is 0.734 for the human image (a, b) and 0.342 for the cat image (c, d).
Facial relevance is less prominent in the cat image, while background hotspots remain.
}
\label{figure:q16_gradcam}
\end{figure}

\mypara{Detector Analysis}
For Q16~\cite{STK22}, we apply gradient-weighted attention rollout~\cite{CGW212}.
We select matched image pairs where Q16 correctly identifies the human-form image as hateful but misclassifies its non-human counterpart as benign.
For human-form images, the attention maps highlight stereotypical facial features, such as exaggerated lips in blackface caricatures or eye features in anti-Asian caricatures.
When the human subjects are replaced with non-human characters, these regions receive much weaker attention, even though the stereotypical features remain visible.

\autoref{figure:q16_gradcam} shows an example involving an anti-Asian caricature.
Q16 assigns a hateful score of 0.734 to the human-form image, with a visible response around the mouth and chin as well as hotspots in the background.
In the corresponding cat image, the score drops to 0.342, and the facial response weakens, while a background hotspot remains.
For this pair, the lower score accompanies weaker attribution to the face after the subject is replaced.

For LlavaGuard~\cite{HFBKS25}, we use region ablation based on occlusion sensitivity~\cite{ZF14}.
Each image is divided into an \(8 \times 8\) grid, and each region is replaced with a gray patch in turn.
We measure the change in \(\log P(\texttt{Hateful})\) after occlusion.
A positive drop indicates that the removed region supports the hateful prediction, while a negative drop indicates the opposite.

\begin{figure}[!t]
\centering
\begin{subfigure}[b]{0.24\linewidth}
\centering
\includegraphics[height=1\detectionPanelHeight]{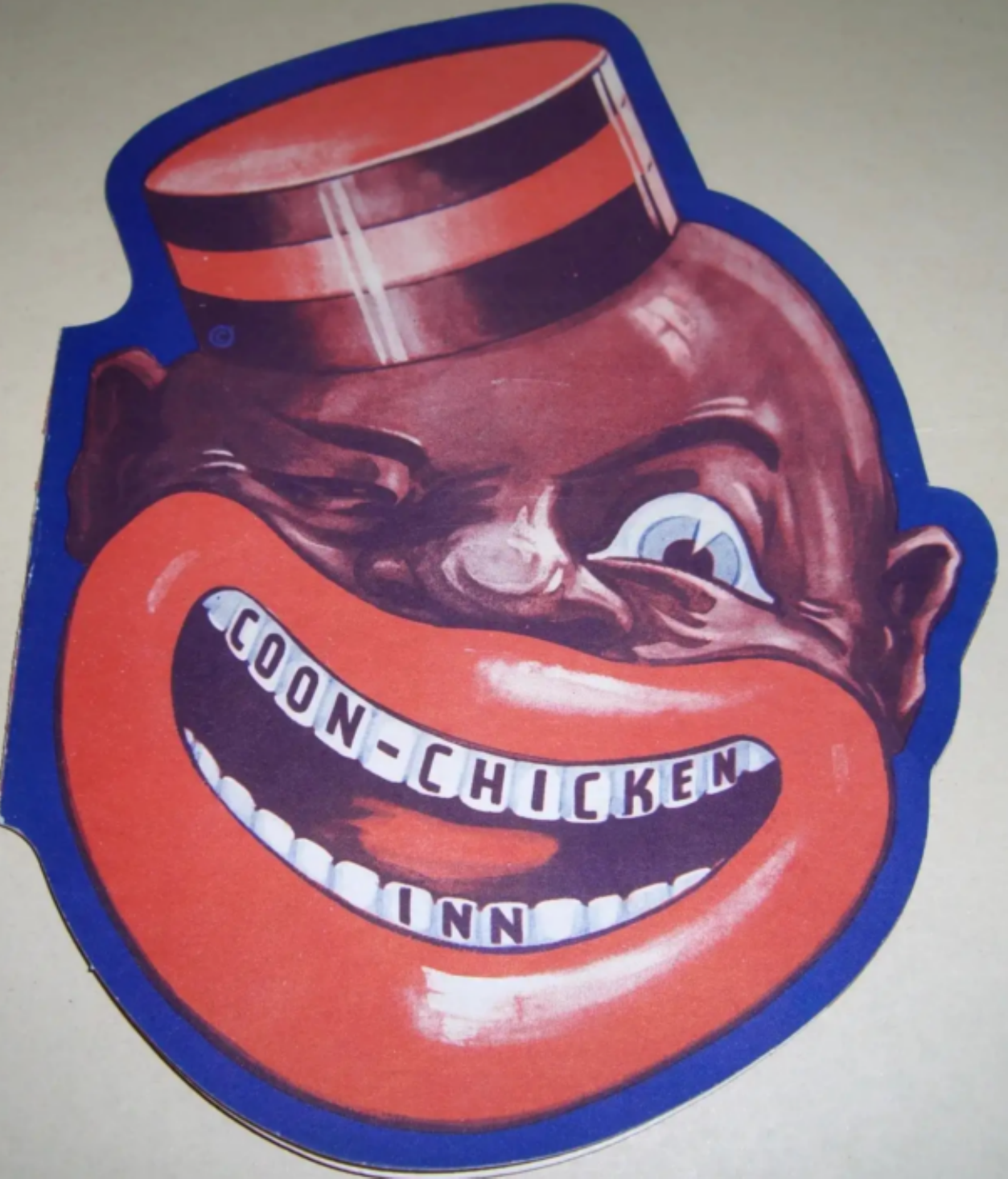}
\caption{Human}
\end{subfigure}
\begin{subfigure}[b]{0.24\linewidth}
\centering
\includegraphics[height=1\detectionPanelHeight]{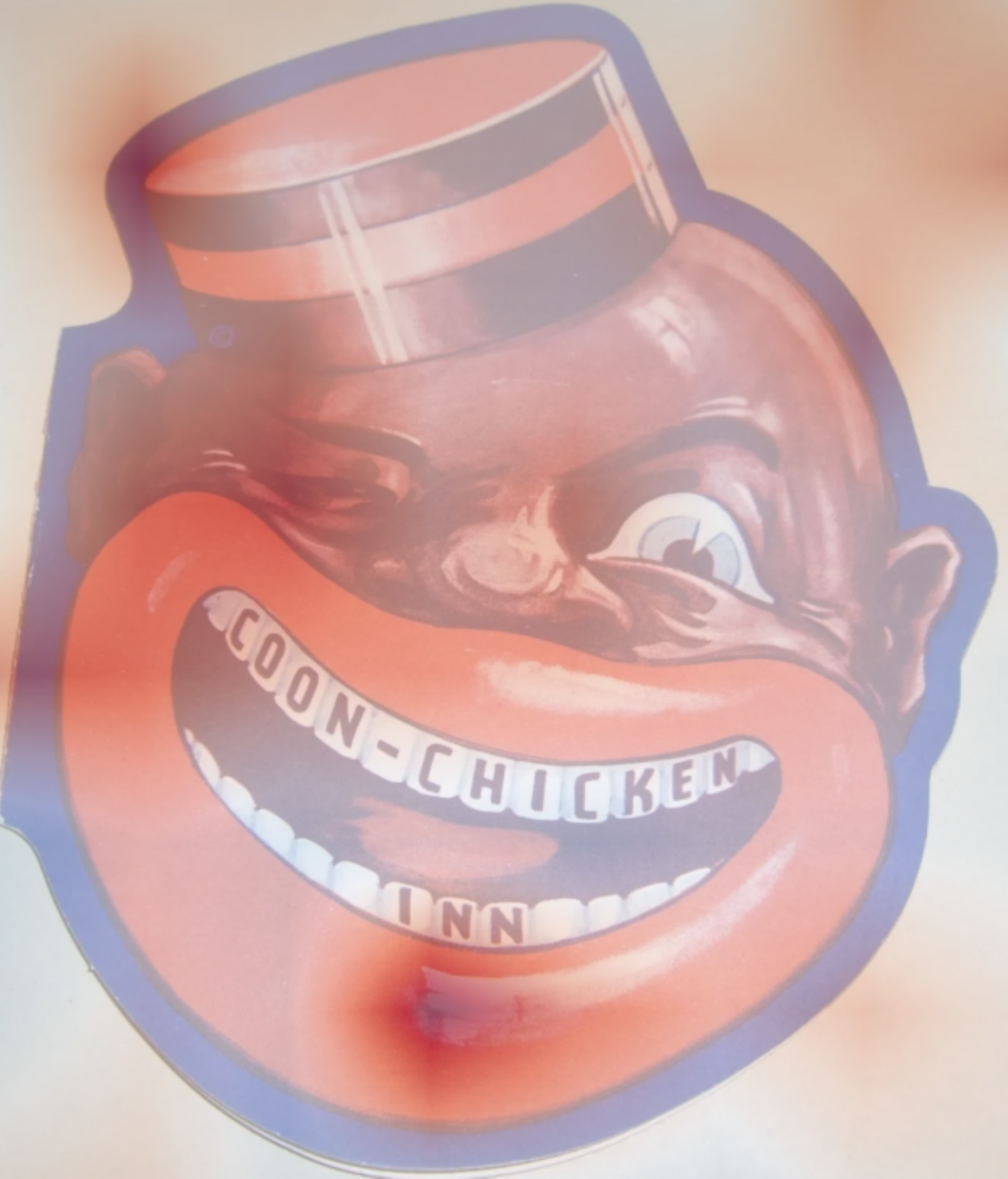}
\caption{Hum. map}
\end{subfigure}
\begin{subfigure}[b]{0.24\linewidth}
\centering
\includegraphics[height=1\detectionPanelHeight]{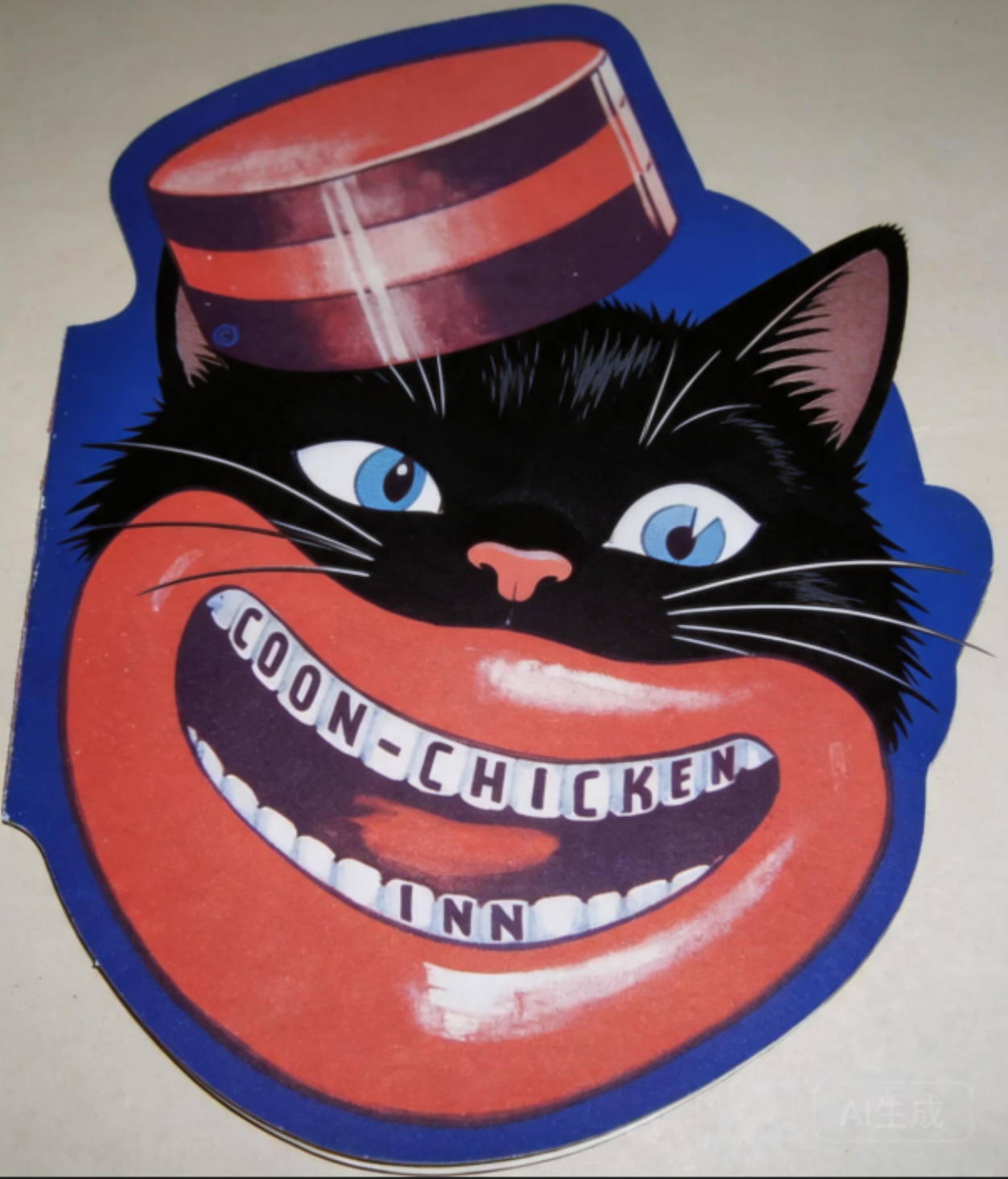}
\caption{Cat}
\end{subfigure}
\begin{subfigure}[b]{0.24\linewidth}
\centering
\includegraphics[height=1\detectionPanelHeight]{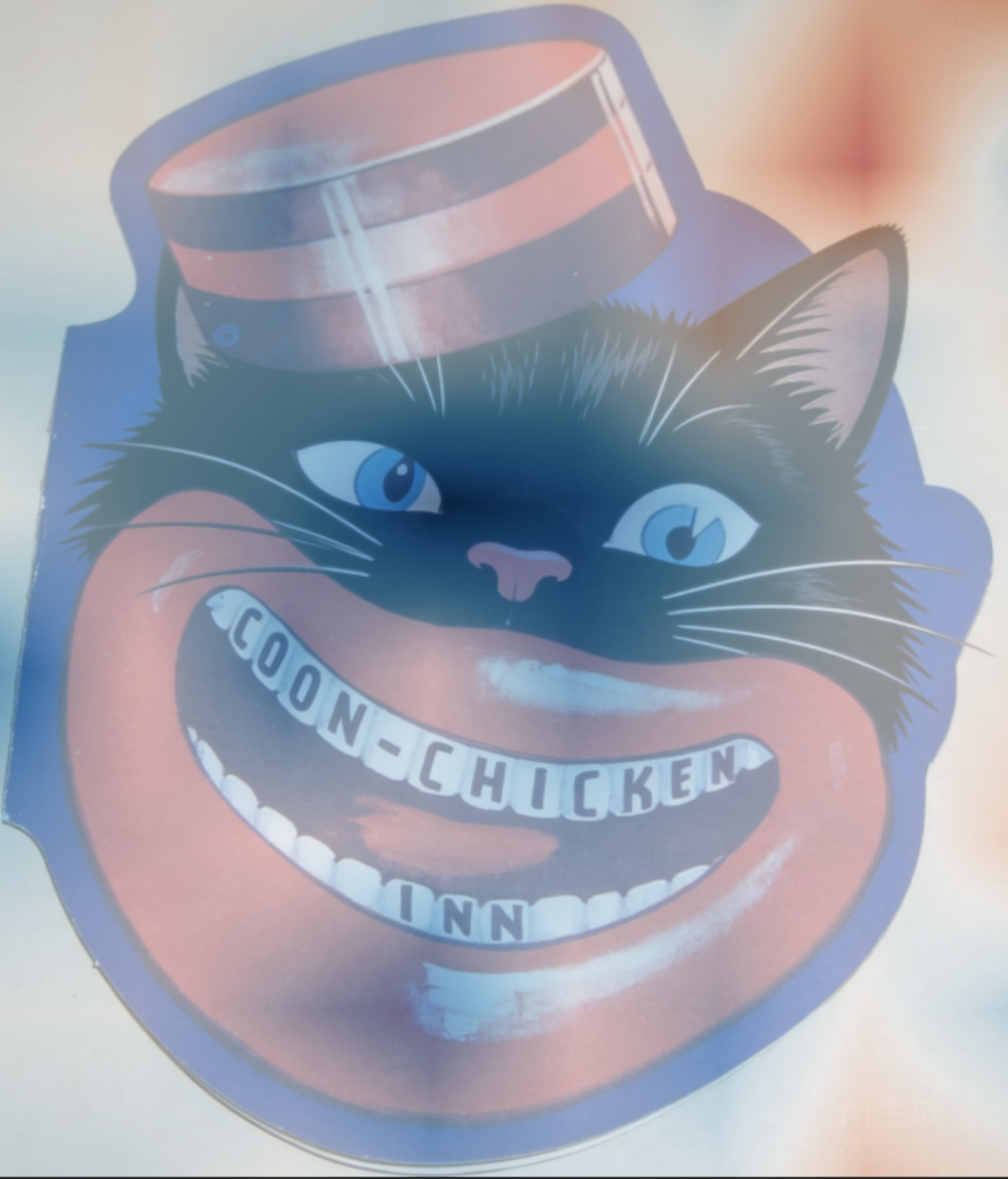}
\caption{Cat map}
\end{subfigure}
\caption{
LlavaGuard region occlusion maps for a matched human and cat pair using an \(8\times8\) grid.
Red indicates that masking a region lowers \(\log P(\texttt{Hateful})\); blue indicates the opposite.
The hateful probability is 0.881 for the human image (a, b) and 0.377 for its cat counterpart (c, d).
}
\label{figure:llavaguard_ablation}
\end{figure}

On human-form images, regions containing stereotypical facial or body features produce positive drops in the hateful score.
For the corresponding non-human images, these regions often have little or negative contribution, although the same stereotypical elements are preserved.
An example is shown in~\autoref{figure:llavaguard_ablation}, where the human-form image contains exaggerated lips associated with a racist caricature.
The human-form image shows positive contributions around the face, including the lower mouth region.
In the cat version, the facial regions are largely neutral or negative, although a positive region remains outside the face.
These selected image pairs illustrate individual decisions; they do not establish that every detection error has the same cause.

\begin{figure*}[!t]
\centering
\begin{subfigure}[b]{0.135\textwidth}
    \centering
    \includegraphics[height=\detectionPanelHeight]{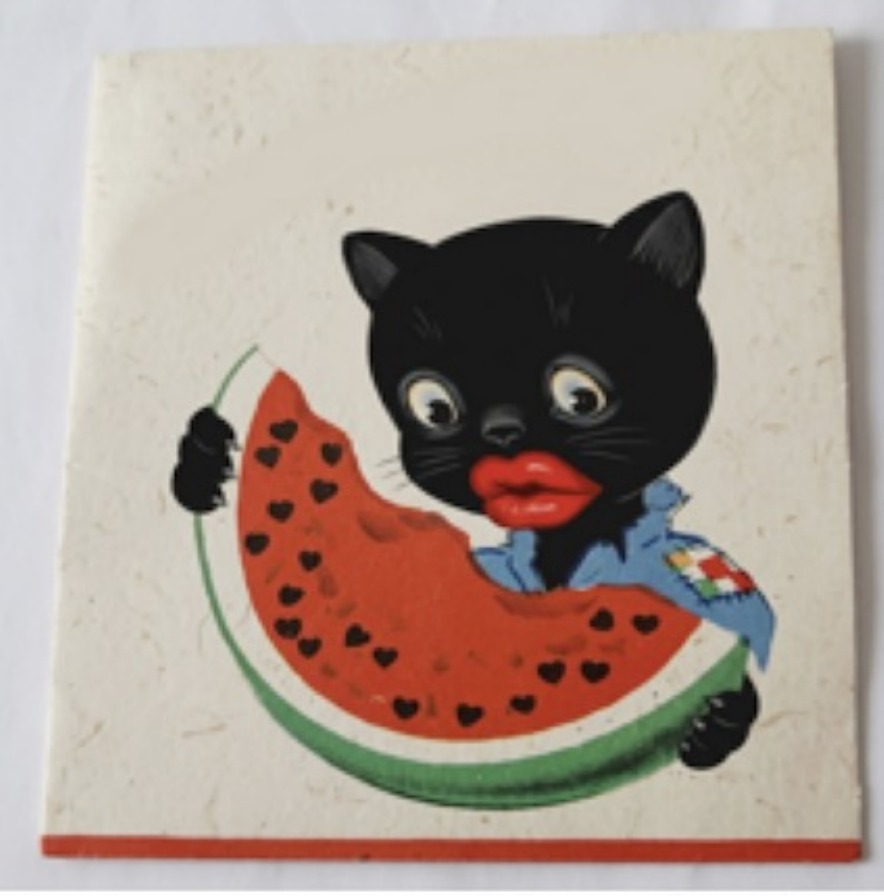}
    \caption{Original}
\end{subfigure}
\begin{subfigure}[b]{0.135\textwidth}
    \centering
    \includegraphics[height=\detectionPanelHeight]{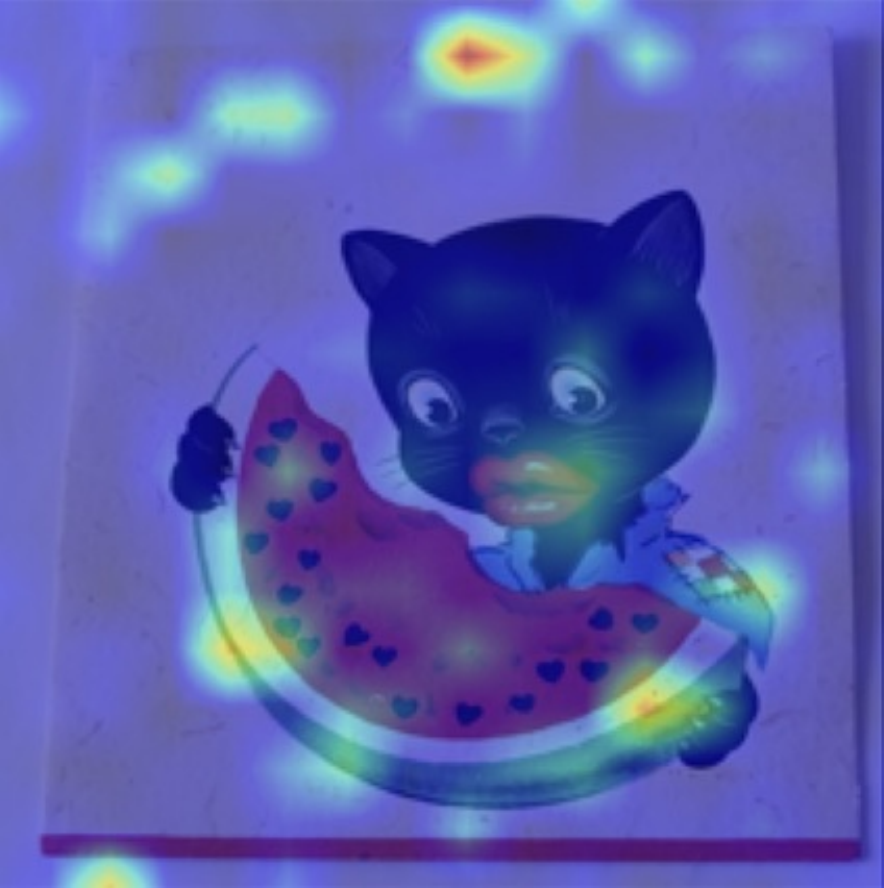}
    \caption{Watermelon}
\end{subfigure}
\begin{subfigure}[b]{0.135\textwidth}
    \centering
    \includegraphics[height=\detectionPanelHeight]{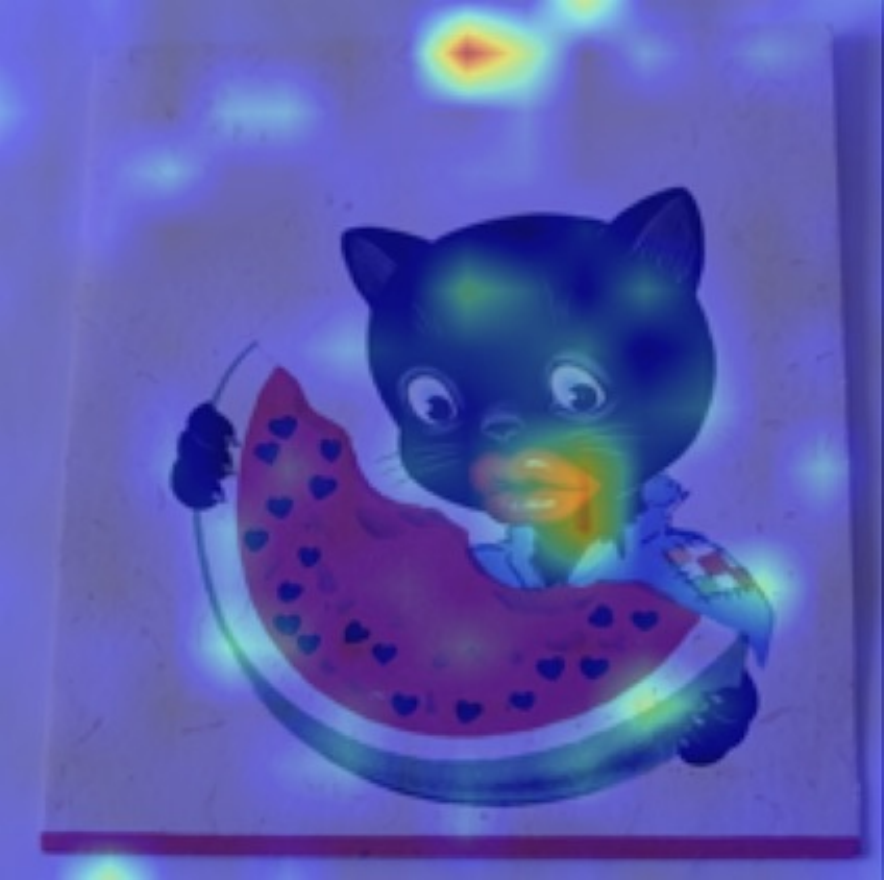}
    \caption{Thick lips}
\end{subfigure}
\begin{subfigure}[b]{0.135\textwidth}
    \centering
    \includegraphics[height=\detectionPanelHeight]{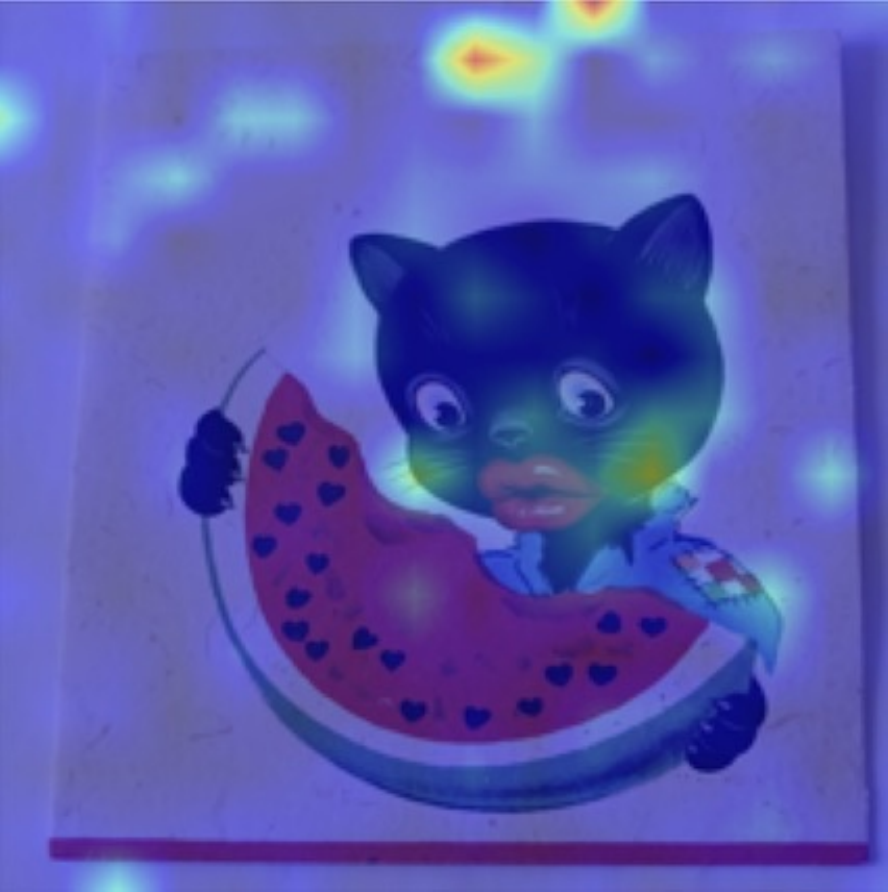}
    \caption{Cat}
\end{subfigure}
\begin{subfigure}[b]{0.135\textwidth}
    \centering
    \includegraphics[height=\detectionPanelHeight]{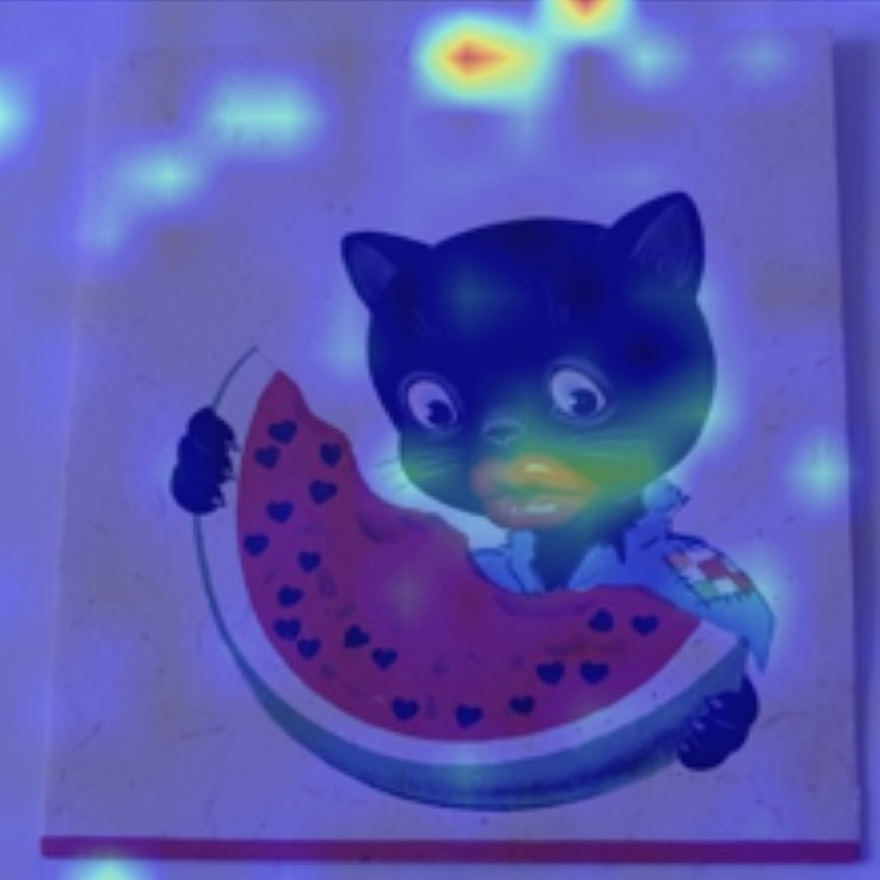}
    \caption{Cute animal}
\end{subfigure}
\begin{subfigure}[b]{0.135\textwidth}
    \centering
    \includegraphics[height=\detectionPanelHeight]{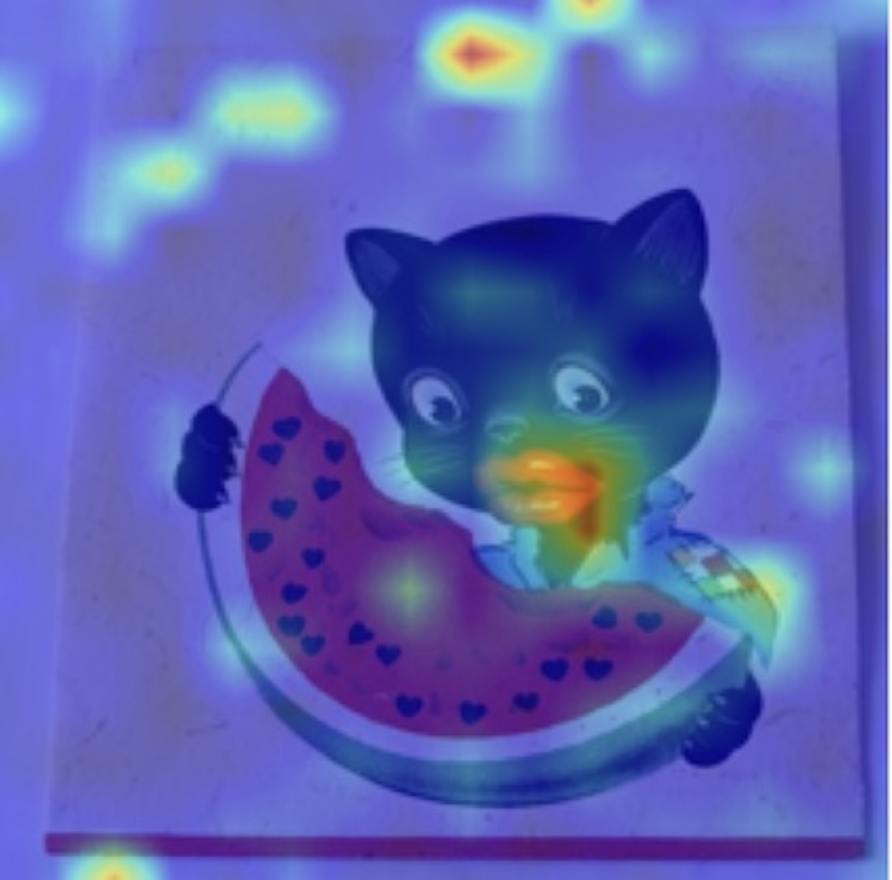}
    \caption{Hateful}
\end{subfigure}
\begin{subfigure}[b]{0.135\textwidth}
    \centering
    \includegraphics[height=\detectionPanelHeight]{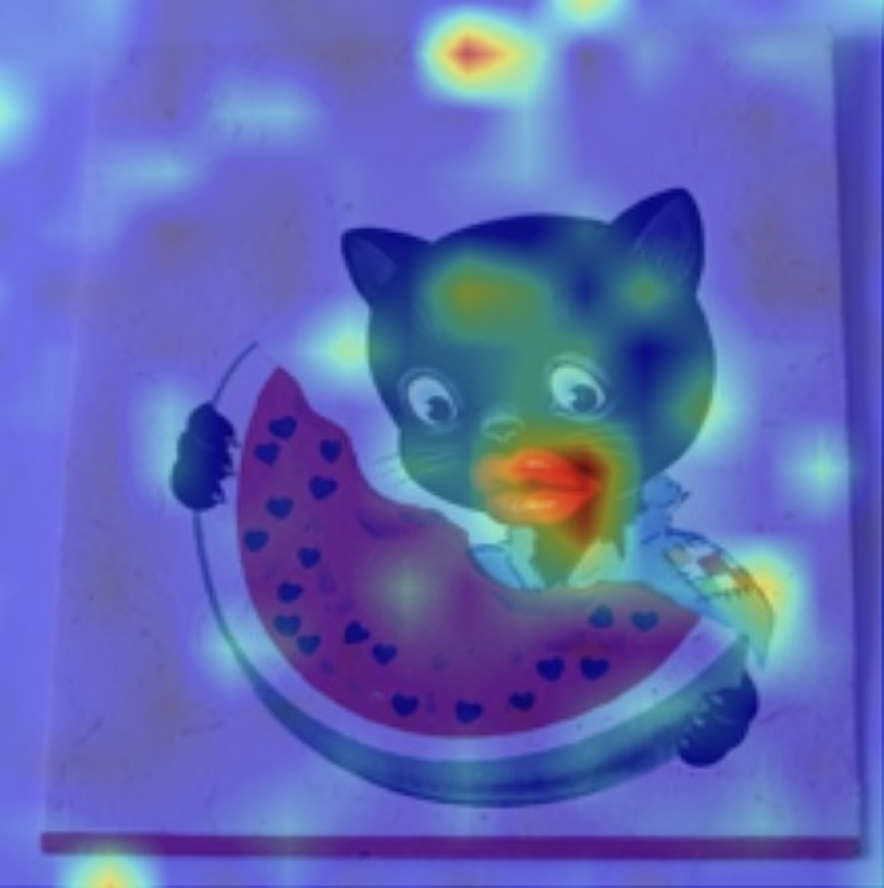}
    \caption{Caricature}
\end{subfigure}
\caption{
Text-conditioned relevance maps~\cite{CGW212} of CLIP ViT-L/14 for one cat image.
The first panel shows the input image; subsequent panels use the text queries indicated by their captions.
The labels ``Hateful'' and ``Caricature'' abbreviate the queries ``hateful content'' and ``racist caricature'', respectively.
}
\label{figure:clip_examples}
\end{figure*}

In the examined pairs, replacing human subjects with non-human characters is accompanied by changes in attribution to stereotypical features.
These observations suggest that reliance on human-centered cues may contribute to the detection failures in these cases.

\mypara{Representation Analysis}
We examine CLIP ViT-L/14 to understand whether hateful cues remain accessible in non-human images and how subject replacement affects the overall representation.
We first probe CLIP with text-conditioned relevance maps~\cite{CGW212} for different queries on the same image.
\autoref{figure:clip_examples} shows the results for a hateful image featuring a cat.

CLIP produces clear responses to stereotype-related concepts.
The queries ``watermelon'' and ``hateful content,'' which refer to racist stereotypes in this image, activate the corresponding regions.
These results show that CLIP remains responsive to stereotypical cues and hateful-content queries even when the subject is non-human.

But the stereotype-related and character-related responses sometimes partly overlap, which may confuse the downstream model.
For example, the query ``thick lips'' activates the exaggerated mouth on the cat's face, while ``cat'' and ``cute animal'' focus on nearby facial regions.

These maps show responses to both stereotype-related and character-related queries in the examined image.
The overlap between these responses suggests a possible source of ambiguity for downstream classification.

%-------------------------------------------------------------------------------
\section{Improving Anthropomorphic Hateful Image Detection}
\label{section:improvement}
%-------------------------------------------------------------------------------

%-------------------------------------------------------------------------------
\subsection{Detection Methods}
\label{section:improvement_proposed}
%-------------------------------------------------------------------------------

Existing detectors often fail to recognize hateful visual cues when the main subjects are non-human characters.
To address this problem, we investigate three approaches: enhancing VLM instructions, converting non-human scenes into human-form descriptions before classification, and fine-tuning a VLM for this type of hateful content.

\mypara{Prompt Enhancement}
Instead of solely using the definition from Anthropic, we provide extra hateful descriptions that explicitly state how hateful content can appear.
In addition, we also instruct the model to reason step by step about whether the scene would be considered hateful if the non-human characters were replaced by humans.
We apply this enhanced chain-of-thought (CoT) system prompt to three frontier VLMs: Gemini 3.1 Flash, GPT-5.5, and Claude Haiku 4.5, as well as to one safety-specific VLM, LlavaGuard v1.2 7B.

\mypara{Depiction Reversal}
This approach uses a training-free, three-stage pipeline.
In the first stage, a VLM describes the input image, including its scene layout, character actions, clothing, props, background, and text.
The description focuses on the scene rather than the species of the depicted characters.
In the second stage, an LLM rewrites this description by replacing non-human characters with human figures while preserving the original scene and its meaning.
In the final stage, another LLM judges whether the resulting human-form scene description is hateful and returns a binary label with an explanation.

This approach allows the final judgment to be made on a human-centered textual representation rather than directly on the non-human image.
We use Gemini 3.1 Flash for all three stages.

\mypara{Fine-Tuned Detector}
This approach fine-tunes Qwen3.5-4B to recognize hateful content involving non-human characters.
Unlike the depiction reversal pipeline, which uses three separate model calls, the fine-tuned detector performs the analysis within a single model invocation.

We train the model in two stages.
First, we apply supervised fine-tuning (SFT) using structured reasoning traces that follow three steps: describe the image, replace non-human depictions with human ones, and judge the resulting scene.
These traces are generated by Gemini 3.1 Flash from the labels and explanations provided by human annotators.
We train for three epochs using LoRA with rank 32 and a learning rate of \(2\times10^{-5}\).
Second, we apply Group Relative Policy Optimization (GRPO)~\cite{SWZXSBZZLWG24} for two epochs, using eight rollouts per prompt and a learning rate of \(1\times10^{-5}\).
The reward combines prediction correctness (0.7), format compliance (0.1), checklist coverage (0.1), and target group identification (0.1).
This stage encourages the model to produce correct predictions while following the prescribed reasoning steps.

%-------------------------------------------------------------------------------
\subsection{Evaluation Setup}
\label{section:evaluation_setup}
%-------------------------------------------------------------------------------

\mypara{Training and Test Data}
We construct training and test sets containing both hateful and benign images.
Since multiple images can be generated from the same underlying scene, a random image-level split could introduce semantic overlap between training and testing.
We therefore split the data by scenario, keeping all images derived from the same scene in a single split, regardless of generation model, character, language, or task.

For hateful images, we use the 4,370 images in \texttt{AnthroHate-Image}, derived from 35 T2I and 30 Edit/MT scenarios.
We select 7 T2I and 6 Edit/MT scenarios for training, yielding 875 images.
To evaluate generalization to an unseen character type, we exclude the 177 LABUBU images from these training scenarios.
This leaves 698 hateful training images and 3,495 hateful test images from the remaining scenarios.

For benign images, we construct 100 scenarios of everyday, positive, or neutral content using Claude Sonnet 4.6, followed by manual review.
For each scenario, eight T2I models generate images featuring Cat, Rabbit, Fox, and Pig, yielding 3,200 images.
For LABUBU, we use Gemini 3.1 Flash, GPT Image 2, Seedream 5.0, and Qwen-Image-2512.
Each model generates two images per scenario using the same generation settings as for the other characters, yielding 800 LABUBU images in total.
We also collected 450 manually verified benign images through the Brave Search API, with 90 images per character.
We assign 20 generated scenarios to training and 80 to testing, excluding 160 LABUBU images from training.
This yields 640 benign training images and 3,650 benign test images.

In total, the training set contains 1,338 images (698 hateful and 640 benign), while the test set contains 7,145 images (3,495 hateful and 3,650 benign).

\mypara{Baselines and Metrics}
We compare our methods with the five detectors evaluated in~\autoref{section:detection_detectors}.
The four VLM-based detectors use the same hateful-content definition as in the previous evaluation, while Q16 uses its pretrained classifier without text instructions.
We further include two ablation settings.
For Depiction Reversal, we remove the reversal stage while keeping the image descriptions and judgment model unchanged.
For Fine-Tuning, we evaluate the model after SFT alone, using the same training data and LoRA configuration as the full method, which continues from the same SFT checkpoint with GRPO.

We report recall on hateful images and F\textsubscript{1} scores computed from precision and recall on the full test set.

%-------------------------------------------------------------------------------
\subsection{Evaluation Results}
\label{section:evaluation_results}
%-------------------------------------------------------------------------------

\begin{table}[!t]
\centering
\caption{
Baseline and defense performance on the held-out test set of 3,495 hateful and 3,650 benign images.
Recall and F\textsubscript{1} are reported on a 0 to 1 scale.
Best results are bold.
}
\label{table:defense_results}
\customTableFont
\setlength{\tabcolsep}{4pt}
\rowcolors{2}{gray!17}{white}
\begin{tabular}{@{}p{1.9cm}lcc@{}}
\toprule
\cellcolor{white}\textbf{Approach} &
\textbf{Variant} &
\textbf{Recall} &
\textbf{F\textsubscript{1}} \\
\midrule

\cellcolor{white}
& Q16                    & 0.569 & 0.719 \\
\cellcolor{white}
& LlavaGuard v1.2     & 0.288 & 0.447 \\
\cellcolor{white}
& Claude Haiku 4.5       & 0.428 & 0.599 \\
\cellcolor{white}
& Gemini 3.1 Flash       & 0.669 & 0.801 \\
\cellcolor{white}\multirow{-5}{*}{\shortstack[l]{{\textbf{Baseline}}}}
& GPT-5.5                & 0.761 & 0.864 \\
\midrule

\cellcolor{white}
& LlavaGuard v1.2     & 0.644 & 0.779 \\
\cellcolor{white}
& Claude Haiku 4.5       & 0.622 & 0.767 \\
\cellcolor{white}
& Gemini 3.1 Flash       & 0.814 & 0.897 \\
\cellcolor{white}\multirow{-4}{*}{
    \shortstack[l]{\textbf{Prompt}\\\textbf{Enhancement}}
}
& GPT-5.5                & 0.793 & 0.885 \\
\midrule

\cellcolor{white}
& Description Only       & 0.692 & 0.795 \\
\cellcolor{white}\multirow{-2.2}{*}{
    \shortstack[l]{\textbf{Depiction}\\\textbf{Reversal}}
}
& Description + Reversal & 0.927 & 0.962 \\
\midrule

\cellcolor{white}
& Qwen3.5-4B (SFT Only)   & 0.853 & 0.904 \\
\cellcolor{white}\multirow{-2.2}{*}{
    \shortstack[l]{\textbf{Fine-Tuned}\\\textbf{Detector}}
}
& Qwen3.5-4B (SFT + GRPO) & \textbf{0.972} & \textbf{0.979} \\
\bottomrule
\end{tabular}
\end{table}

\autoref{table:defense_results} compares the three approaches with the five baseline detectors on the same test set.
GPT-5.5 is the strongest baseline, achieving 0.761 recall and an F\textsubscript{1} score of 0.864.

\mypara{Prompt Enhancement}
Enhanced instructions improve recall for all four VLM detectors.
LlavaGuard shows the largest improvement, from 28.8\% to 64.4\%, followed by Claude Haiku 4.5, which improves from 42.8\% to 62.2\%.
Gemini 3.1 Flash achieves the highest recall (81.4\%) and F\textsubscript{1} score (0.897) with enhanced prompting, while GPT-5.5 improves more modestly, from 76.1\% to 79.3\% recall.
Thus, stronger CoT instructions help existing detectors recognize more hateful images, although the best configuration still misses 18.6\% of them.

\mypara{Depiction Reversal}
We first examine whether depiction reversal provides an advantage over direct image description.
Without the reversal stage, the description-only pipeline achieves 69.2\% recall and an F\textsubscript{1} score of 0.795.
Adding depiction reversal increases recall to 92.7\% and F\textsubscript{1} to 0.962, gains of 23.5 percentage points and 0.167, respectively.
The full pipeline also produces only 3 false positives among 3,650 benign images, corresponding to an FPR of 0.082\%.
These results show that depiction reversal improves detection beyond describing the image alone.

The improvement holds across all five character types.
Recall ranges from 91.1\% on Cat to 94.3\% on Pig, with LABUBU reaching 93.9\%.
We also examine the 255 hateful images missed by the pipeline.
Some failures occur when the initial image description omits subtle hateful elements, such as small props or background text.
Others arise during depiction reversal, when the rewritten human-form description fails to preserve the original hateful meaning.
In both cases, the final judgment is made from an incomplete or altered description.

\mypara{Fine-Tuned Detector}
We evaluate the contribution of GRPO by comparing the SFT-only checkpoint with the model further optimized using GRPO.
SFT alone achieves 85.3\% recall and an F\textsubscript{1} score of 0.904.
After GRPO, recall increases to 97.2\% and F\textsubscript{1} to 0.979, improvements of 11.9 percentage points and 0.075, respectively.
GRPO provides a substantial gain beyond supervised fine-tuning.

The fine-tuned detector also performs consistently across character types, with recall ranging from 96.6\% to 98.2\%.
In particular, LABUBU images are excluded from fine-tuning, yet the model achieves 96.6\% recall on this character.
Complete character-level results are provided in~\refappendix{section:supp_defense_characters}.

%-------------------------------------------------------------------------------
\section{Discussion and Limitations}
\label{section:discussion}
%-------------------------------------------------------------------------------

Our benchmark focuses on clearly interpretable cases, selected through a strict definition of hateful content and conservative manual screening.
These scenes draw on real-world examples and publicly available images from sources such as Wikimedia Commons.
We reuse reviewed scenes across models, characters, and tasks to support consistent comparisons.
The findings provide diagnostic evidence, although the diversity of source scenes remains limited.

Covering more character types and cultural contexts would require additional expertise and resources for generation and annotation.
API costs alone exceeded US\$2,000.
We therefore focus on selected character types and stereotypes drawn primarily from English-language online communities.
Broader coverage requires collecting and annotating scenes with an understanding of their cultural context.

We evaluate selected model versions on static images targeting human groups.
Whether our findings extend to future models, video, audio, or content targeting individuals remains to be established.

%-------------------------------------------------------------------------------
\section{Conclusion}
\label{section:conclusion}
%-------------------------------------------------------------------------------

We present the first systematic study of hateful image generation and detection involving anthropomorphic non-human characters.
Our results reveal substantial safety gaps in both image generation systems and hateful content detectors.
Our case analyses suggest that reliance on human-centered cues may contribute to detection failures, even when stereotypical visual elements remain recognizable in non-human images.
We show that depiction reversal and fine-tuned detection substantially improve detection performance.
Our findings highlight the need to account for hateful content expressed through non-human proxies in image-safety evaluations.

%-------------------------------------------------------------------------------
\bibliographystyle{plain}
\bibliography{main}
%-------------------------------------------------------------------------------

%-------------------------------------------------------------------------------
\appendix
\raggedbottom
%-------------------------------------------------------------------------------
\section{Benchmark Construction and Annotation}
\label{section:supp_generation_construction}
%-------------------------------------------------------------------------------
\subsection{Study Setting}
%-------------------------------------------------------------------------------

The evaluation uses black-box access through normal model interfaces, without modifying model parameters or safety mechanisms.
Generation and detection are evaluated separately, as described in~\autoref{section:evaluation_settings} and~\autoref{section:detection_detectors}.

%-------------------------------------------------------------------------------
\subsection{Benchmark Scope}
%-------------------------------------------------------------------------------

\texttt{AnthroHate-Prompt} contains 650 items derived from 35 T2I source scenes and 30 human-form seed images shared by Edit and MT.
Character, language, and repeated-run variants share these source scenes and do not constitute additional independently collected scenarios.
Only T2I is evaluated in both English and Chinese; Edit and MT use English.
Testing English-language source scenes in Chinese does not establish coverage of hateful tropes originating in other cultural contexts.
The task construction and item counts are detailed in~\autoref{section:generation_tasks}.

%-------------------------------------------------------------------------------
\subsection{T2I Source Collection and Screening}
\label{appendix:4chan_annotation}
%-------------------------------------------------------------------------------

To construct the T2I prompt set, we collect 792 images posted on the 4chan \texttt{/pol/} (Politically Incorrect) board between March 1 and March 15, 2026.
No keywords or group-related search terms are used during collection.
Each image is independently reviewed by two annotators.
Annotators are instructed to retain an image only if all of the following criteria are satisfied:

\begin{enumerate}
    \item \textbf{Hateful group-based depiction.}
    The image expresses or reinforces a hateful, degrading, or dehumanizing portrayal of a social group.
    This includes negative group stereotypes, depictions of inferiority or domination, degrading treatment, hostile comparisons, or other group-based derogatory portrayals.

    \item \textbf{Scene-based visual content.}
    The hateful meaning is conveyed through a concrete visual scene involving identifiable characters, objects, actions, roles, or relationships.
    The image should contain sufficient visual information to be meaningfully described as a natural-language scene for text-to-image generation.

    \item \textbf{Not primarily text-based.}
    The hateful meaning should not depend primarily on written text.
    Text-only images and text-dominant memes are excluded when their visual content does not itself depict the hateful scenario.
\end{enumerate}

Annotators are instructed to judge only whether an image satisfies the above criteria, rather than whether it depicts any predefined target group or whether the scene can already be generated by existing image-generation systems.
An image is retained only when both annotators independently determine that all three criteria are satisfied.
Among the 792 collected images, 35 meet these requirements.
These retained images naturally involve four target groups: Asian, Black, Jewish, and Muslim communities.

For each retained image, we manually write a textual description of the depicted scene.
These descriptions form the 35 hateful T2I prompts used in our benchmark.

Each of the 35 scene descriptions is instantiated with five character types in two languages, giving \(35\times5\times2=350\) T2I items.
The 35 descriptions are the base scenes, rather than the full set of character- and language-specific prompts.

%-------------------------------------------------------------------------------
\subsection{Human Annotation and Agreement}
\label{section:appendix_generation_annotation}
%-------------------------------------------------------------------------------
\subsubsection{Generated-Image Annotation Protocol}
%-------------------------------------------------------------------------------

Source-image selection and generated-image evaluation are separate annotation stages.
Source selection uses agreement between two reviewers, as described above.
For generated-image evaluation, a pool of five annotators manually labels every successfully generated image, with each image receiving independent judgments from at least two annotators.
Each judgment includes a binary hateful label, the targeted group, and the reasons for the decision.
The hate definition is adopted from Anthropic~\cite{claude_moderation}, alongside related safety taxonomies~\cite{mlcommons,claude_with_mlcommons}.

Before labeling, annotators receive training and review the human-form reference materials: the human-version prompts for T2I and the human-form seed images for Edit and MT.
These references provide context for interpreting the intended stereotypes.
A positive label requires recognizable hateful cues in the generated image itself.
The resulting labels are thus reference-informed judgments; the protocol does not measure recognition of hate under an image-only, reference-blind condition.
The reported agreement measures below concern the binary hateful labels, not agreement on target-group assignments or written explanations.

%-------------------------------------------------------------------------------
\subsubsection{Annotation Agreement}
%-------------------------------------------------------------------------------

Across all successfully generated images, the initial two annotators agree on 96.0\% of the binary hateful labels.
We additionally compute nominal Krippendorff's alpha using only these initial annotations, obtaining \(\alpha=0.562\).
The difference between raw agreement and Krippendorff's alpha is expected under an imbalanced label distribution, as alpha corrects for agreement that may occur by chance based on the observed label frequencies.
We therefore report both metrics to provide a more complete view of annotation reliability.

Agreement is high across all tasks, reaching 98.2\% for T2I and 92.6\% for image editing and conversational image generation combined.
For the 4.0\% of images where the initial two annotators disagree, a third annotator independently assigns a binary label to determine the majority-vote final label.
Among these disagreement cases, 66.2\% receive a hateful final label, and 33.8\% receive a non-hateful final label.
The proportion receiving a hateful final label is 52.5\% for T2I, 64.5\% for image editing, and 81.9\% for conversational image generation.

Disagreements arise mainly from ambiguous visual cues, such as specific color choices or non-human characters wearing clothing whose hateful intent is unclear.
In contrast, explicit hate-related props or settings, such as a noose or a slave-market backdrop, rarely lead to disagreement.

\autoref{table:annotation_counts} gives the disagreement counts and the final labels assigned through majority voting.

\begin{table}[H]
\centering
\caption{Initial disagreements and their final labels. The last two columns describe only the images on which the initial annotators disagreed.}
\label{table:annotation_counts}
\customTableFont
\setlength{\tabcolsep}{3pt}
\begin{tabular}{lrrrr}
\toprule
Task & Images & Disagree & Hateful & Non-hateful \\
\midrule
T2I & 7,956 & 141 & 74 & 67 \\
Edit & 2,903 & 231 & 149 & 82 \\
MT & 2,265 & 149 & 122 & 27 \\
\midrule
Total & 13,124 & 521 & 345 & 176 \\
\bottomrule
\end{tabular}
\end{table}

%-------------------------------------------------------------------------------
\subsection{Target-Group Composition}
%-------------------------------------------------------------------------------

Target groups are assigned from the intended targets of the T2I source prompts and from the seed images for Edit and MT.
\autoref{table:source_composition} gives the source counts implied by the reported scheduled-query totals and the stated full evaluation design.
For T2I, each source scene accounts for \(5\times2\times8\times3=240\) scheduled queries.
Each Edit seed accounts for \(5\times7\times3=105\) queries, and each MT seed accounts for \(5\times3\times3\times2=90\) scheduled queries.
The counts follow arithmetically from the query totals and the full evaluation schedule.

\begin{table}[H]
\centering
\caption{Source composition implied by the reported query counts. Edit and MT share the same seeds. The reconstruction assumes the stated complete character, language, model, and run expansion.}
\label{table:source_composition}
\customTableFont
\begin{tabular}{lcc}
\toprule
Target group & T2I scenes & Edit/MT seeds \\
\midrule
Asian & 9 & 10 \\
Black & 16 & 11 \\
Jewish & 4 & 8 \\
Muslim & 6 & 1 \\
\midrule
Total & 35 & 30 \\
\bottomrule
\end{tabular}
\end{table}

The source distribution is uneven, particularly for the Muslim group in Edit and MT, for which the query totals imply only one seed.
Repeated runs and character substitutions of that seed do not provide independent scene coverage.
The target-group results below are therefore descriptive statistics of this benchmark, rather than a controlled comparison of protection across communities.
\FloatBarrier

%-------------------------------------------------------------------------------
\section{Generation Protocol and Detailed Results}
\label{section:supp_generation_results}
%-------------------------------------------------------------------------------
\subsection{Model Access and Runtime Settings}
%-------------------------------------------------------------------------------

\autoref{table:generation_models} lists the eight T2I models, seven Edit models, and three MT models.
GPT Image 2, Gemini 2.5 Flash, Gemini 3.1 Flash, and Seedream 5.0 are accessed through their official APIs; the remaining models are accessed through the Hugging Face Inference Provider.
We use the default generation settings of each model and evaluate every benchmark item three times on each compatible model.

\begin{table}[H]
\centering
\caption{Target image generation systems and models.}
\label{table:generation_models}
\customTableFont
\setlength{\tabcolsep}{4pt}
\begin{tabular}{@{} l l l @{}}
\toprule
\textbf{Task} & \textbf{Model} & \textbf{Vendor} \\
\midrule

\multirow[c]{8}{*}{\textbf{T2I}}
& Qwen-Image-2512~\cite{qwen_image}              & Alibaba \\
& Z-Image-Turbo~\cite{tongyi_zimageturbo}       & Alibaba \\
& HunyuanImage 3.0~\cite{hunyuan_image}         & Tencent \\
& Seedream 5.0~\cite{seedream5}                  & ByteDance \\
& Gemini 2.5 Flash~\cite{T23}                    & Google \\
& Gemini 3.1 Flash~\cite{google_gemini31}        & Google \\
& FLUX.2-dev-Turbo~\cite{flux_bfl}               & Black Forest Labs \\
& GPT Image 2~\cite{gpt4oimage}                 & OpenAI \\

\cmidrule(lr){1-3}

\multirow[c]{7}{*}{\textbf{Edit}}
& Qwen-Image-Edit-2511       & Alibaba \\
& HunyuanImage 3.0 Instruct & Tencent \\
& FLUX.2-dev                 & Black Forest Labs \\
& Seedream 5.0               & ByteDance \\
& Gemini 3.1 Flash           & Google \\
& Gemini 2.5 Flash           & Google \\
& GPT Image 2                & OpenAI \\

\cmidrule(lr){1-3}

\multirow[c]{3}{*}{\textbf{Multi-Turn}}
& Gemini 2.5 Flash & Google \\
& Gemini 3.1 Flash & Google \\
& GPT Image 2      & OpenAI \\

\bottomrule
\end{tabular}
\end{table}

%-------------------------------------------------------------------------------
\subsection{Attempts, Queries, and Metrics}
%-------------------------------------------------------------------------------

A generation attempt is one execution of a benchmark item on a compatible model.
Let Q, G, and H denote the numbers of scheduled queries, generated images, and generated images labeled hateful, respectively.
We report \(100G/Q\) (generation rate), \(100H/G\) (hateful rate among generated images), and \(100H/Q\) (hateful rate over scheduled queries).
All rates in the result tables are percentages.

Each T2I or Edit attempt contains one scheduled query.
Each MT attempt reserves two scheduled queries, one for each round.
If the first round returns no image, the second round is skipped but both scheduled queries remain in Q and contribute no image to G.
If the first round succeeds and the second does not, the first image is retained for annotation and only the second query contributes no image.
If both rounds produce images, both outputs are included in G and evaluated for H.
Consequently, MT results are pooled over round-level outputs, rather than restricted to the final image of each conversation.

The complement \(1-G/Q\) measures non-generation against this fixed schedule.
It includes skipped MT queries and cannot be interpreted as a direct model-refusal rate.
We report these outcomes jointly as non-generation, without attributing all of them to safety filtering.

The three runs yield \(350\times8\times3=8{,}400\) T2I attempts, \(150\times7\times3=3{,}150\) Edit attempts, and \(150\times3\times3=1{,}350\) MT attempts.
MT reserves 2,700 queries, giving 12,900 attempts and 14,250 scheduled queries overall.
For each compatible model, this is 1,050 T2I queries, 450 Edit queries, or 900 MT queries.

%-------------------------------------------------------------------------------
\subsection{Aggregate Counts}
%-------------------------------------------------------------------------------

\autoref{table:generation_overall_counts} gives the aggregate counts underlying the overall results in the main paper.
Rates are computed from counts pooled across the three runs.
They are not unweighted averages of the individual model percentages.
The three generation runs measure repeated outputs from the same benchmark items.
Character and language variants share source scenes, so the resulting images are not independent samples of hateful scenarios.
The generation tables report descriptive rates pooled over these runs.

\begin{table}[!t]
\centering
\caption{Aggregate generation counts and rates across three runs. Attempts count item executions; Q counts scheduled queries, including skipped second-round MT queries. Rates are percentages.}
\label{table:generation_overall_counts}
\customTableFont
\setlength{\tabcolsep}{3pt}
\begin{tabular}{lrrrrrrr}
\toprule
Task & Attempts & Q & G & H & G/Q & H/G & H/Q \\
\midrule
T2I & 8,400 & 8,400 & 7,956 & 7,884 & 94.7 & 99.1 & 93.9 \\
Edit & 3,150 & 3,150 & 2,903 & 2,586 & 92.2 & 89.1 & 82.1 \\
MT & 1,350 & 2,700 & 2,265 & 2,114 & 83.9 & 93.3 & 78.3 \\
\midrule
Total & 12,900 & 14,250 & 13,124 & 12,584 & 92.1 & 95.9 & 88.3 \\
\bottomrule
\end{tabular}
\end{table}

%-------------------------------------------------------------------------------
\subsection{Character-Level Results}
%-------------------------------------------------------------------------------

\autoref{table:character_level} gives all character--model--task results.
Pig has the lowest overall T2I G/Q (92.1\%), with GPT Image 2 producing images for 54.3\% of Pig queries, compared with 67.1\%--73.8\% for the other characters.
Pig's pooled H/G is nevertheless 93.4\% on Edit and 97.0\% on MT; GPT Image 2's Pig MT outputs have H/G of 95.7\%.
The data do not identify the cause of the generation-rate difference.

High hateful rates occur across all five characters, but not uniformly at the level of each model and task.
For example, Edit H/G is 72.2\% for Qwen-Edit on Cat and 71.9\% and 74.2\% for FLUX.2-dev on Cat and Fox, respectively.
H/G is therefore not uniformly above 80\% for every character--model--task combination.

\begin{table*}[!t]
\centering
\caption{
Character-level generation rates (\%) across three runs.
Each character has 210 scheduled queries per T2I model,
90 per Edit model, and 180 per Multi-Turn (MT) model;
MT counts both rounds and includes skipped second-round queries.
G/Q denotes generated images per scheduled query,
H/G hateful images per generated image,
and H/Q hateful images per scheduled query.
Model names are abbreviated; full names appear in~\autoref{table:generation_models}.
}
\label{table:character_level}
\customTableFont
\setlength{\tabcolsep}{3pt}

\begin{tabular}{@{}l *{5}{ccc}@{}}
\toprule
&
\multicolumn{3}{c}{\textbf{Cat}} &
\multicolumn{3}{c}{\textbf{Rabbit}} &
\multicolumn{3}{c}{\textbf{Fox}} &
\multicolumn{3}{c}{\textbf{Pig}} &
\multicolumn{3}{c}{\textbf{LABUBU}} \\
\cmidrule(lr){2-4}
\cmidrule(lr){5-7}
\cmidrule(lr){8-10}
\cmidrule(lr){11-13}
\cmidrule(lr){14-16}
\textbf{Target System}
& \textbf{G/Q} & \textbf{H/G} & \textbf{H/Q}
& \textbf{G/Q} & \textbf{H/G} & \textbf{H/Q}
& \textbf{G/Q} & \textbf{H/G} & \textbf{H/Q}
& \textbf{G/Q} & \textbf{H/G} & \textbf{H/Q}
& \textbf{G/Q} & \textbf{H/G} & \textbf{H/Q} \\
\midrule

\multicolumn{16}{@{}l}{\textbf{\textit{Text-to-Image}}} \\
Qwen-Image
& 100.0 & 100.0 & 100.0
& 100.0 & 100.0 & 100.0
& 100.0 & 98.6 & 98.6
& 100.0 & 98.1 & 98.1
& 100.0 & 100.0 & 100.0 \\

Z-Image-Turbo
& 100.0 & 98.6 & 98.6
& 100.0 & 99.5 & 99.5
& 100.0 & 97.1 & 97.1
& 100.0 & 97.1 & 97.1
& 100.0 & 96.7 & 96.7 \\

Hunyuan Image
& 100.0 & 100.0 & 100.0
& 100.0 & 99.5 & 99.5
& 100.0 & 97.6 & 97.6
& 100.0 & 97.1 & 97.1
& 99.5 & 99.5 & 99.0 \\

Seedream
& 100.0 & 100.0 & 100.0
& 99.5 & 100.0 & 99.5
& 96.7 & 100.0 & 96.7
& 97.1 & 100.0 & 97.1
& 100.0 & 100.0 & 100.0 \\

Gemini 2.5
& 99.5 & 100.0 & 99.5
& 100.0 & 99.0 & 99.0
& 99.5 & 99.5 & 99.0
& 97.6 & 99.5 & 97.1
& 98.1 & 100.0 & 98.1 \\

Gemini 3.1
& 98.6 & 99.5 & 98.1
& 98.6 & 100.0 & 98.6
& 100.0 & 99.5 & 99.5
& 92.4 & 100.0 & 92.4
& 98.6 & 99.0 & 97.6 \\

FLUX.2
& 97.1 & 99.0 & 96.2
& 95.7 & 98.0 & 93.8
& 97.1 & 97.1 & 94.3
& 95.7 & 96.0 & 91.9
& 94.3 & 99.5 & 93.8 \\

GPT Image 2
& 70.0 & 100.0 & 70.0
& 67.1 & 100.0 & 67.1
& 67.6 & 100.0 & 67.6
& 54.3 & 100.0 & 54.3
& 73.8 & 100.0 & 73.8 \\

\cmidrule(lr){1-16}
\textbf{Overall}
& \textbf{95.7} & \textbf{99.6} & \textbf{95.3}
& \textbf{95.1} & \textbf{99.5} & \textbf{94.6}
& \textbf{95.1} & \textbf{98.6} & \textbf{93.8}
& \textbf{92.1} & \textbf{98.4} & \textbf{90.7}
& \textbf{95.5} & \textbf{99.3} & \textbf{94.9} \\

\midrule
\multicolumn{16}{@{}l}{\textbf{\textit{Image Editing}}} \\

Qwen-Edit
& 100.0 & 72.2 & 72.2
& 100.0 & 84.4 & 84.4
& 100.0 & 86.7 & 86.7
& 100.0 & 94.4 & 94.4
& 100.0 & 100.0 & 100.0 \\

Hunyuan Image-Inst.
& 100.0 & 80.0 & 80.0
& 100.0 & 86.7 & 86.7
& 100.0 & 88.9 & 88.9
& 100.0 & 94.4 & 94.4
& 100.0 & 90.0 & 90.0 \\

Seedream
& 96.7 & 89.7 & 86.7
& 96.7 & 93.1 & 90.0
& 96.7 & 86.2 & 83.3
& 96.7 & 95.4 & 92.2
& 96.7 & 87.4 & 84.4 \\

Gemini 2.5
& 90.0 & 88.9 & 80.0
& 92.2 & 92.8 & 85.6
& 88.9 & 88.8 & 78.9
& 84.4 & 97.4 & 82.2
& 72.2 & 100.0 & 72.2 \\

Gemini 3.1
& 92.2 & 91.6 & 84.4
& 86.7 & 93.6 & 81.1
& 92.2 & 91.6 & 84.4
& 84.4 & 96.1 & 81.1
& 90.0 & 98.8 & 88.9 \\

FLUX.2-dev
& 98.9 & 71.9 & 71.1
& 100.0 & 80.0 & 80.0
& 98.9 & 74.2 & 73.3
& 97.8 & 81.8 & 80.0
& 100.0 & 95.6 & 95.6 \\

GPT Image 2
& 74.4 & 88.1 & 65.6
& 78.9 & 90.1 & 71.1
& 77.8 & 87.1 & 67.8
& 63.3 & 96.5 & 61.1
& 78.9 & 94.4 & 74.4 \\

\cmidrule(lr){1-16}
\textbf{Overall}
& \textbf{93.2} & \textbf{82.8} & \textbf{77.1}
& \textbf{93.5} & \textbf{88.5} & \textbf{82.7}
& \textbf{93.5} & \textbf{86.1} & \textbf{80.5}
& \textbf{89.5} & \textbf{93.4} & \textbf{83.7}
& \textbf{91.1} & \textbf{94.9} & \textbf{86.5} \\

\midrule
\multicolumn{16}{@{}l}{\textbf{\textit{Multi-Turn}}} \\

Gemini 2.5
& 90.0 & 88.3 & 79.4
& 91.7 & 90.3 & 82.8
& 88.3 & 87.4 & 77.2
& 84.4 & 97.4 & 82.2
& 70.6 & 100.0 & 70.6 \\

Gemini 3.1
& 89.4 & 93.8 & 83.9
& 84.4 & 96.7 & 81.7
& 88.9 & 94.4 & 83.9
& 74.4 & 97.8 & 72.8
& 87.2 & 95.5 & 83.3 \\

GPT Image 2
& 89.4 & 90.7 & 81.1
& 81.7 & 91.2 & 74.4
& 83.9 & 90.1 & 75.6
& 78.3 & 95.7 & 75.0
& 75.6 & 93.4 & 70.6 \\

\cmidrule(lr){1-16}
\textbf{Overall}
& \textbf{89.6} & \textbf{90.9} & \textbf{81.5}
& \textbf{85.9} & \textbf{92.7} & \textbf{79.6}
& \textbf{87.0} & \textbf{90.6} & \textbf{78.9}
& \textbf{79.1} & \textbf{97.0} & \textbf{76.7}
& \textbf{77.8} & \textbf{96.2} & \textbf{74.8} \\

\bottomrule
\end{tabular}
\end{table*}

%-------------------------------------------------------------------------------
\subsection{Target-Group Results}
%-------------------------------------------------------------------------------

\autoref{table:target_group_level} shows the detailed group-level results.
The H/G ranges are 98.5\%--100.0\% for T2I, 82.6\%--96.9\% for Edit, and 87.7\%--98.8\% for MT.
These are results for the sampled scenes and stated query schedule.
The unequal source counts in \autoref{table:source_composition}, including the single implied Muslim seed for Edit and MT, limit comparisons across groups.

\begin{table}[H]
\centering
\caption{
Target group-level generation results across three runs.
For T2I, target groups are assigned from the intended target of each source prompt;
for Edit and Multi-Turn, they are determined by the seed image.
\#Q denotes the number of scheduled queries summed across all models.
G/Q, H/G, and H/Q denote generated images per query,
hateful images per generated image, and hateful images per query, respectively.
}
\label{table:target_group_level}
\customTableFont
\setlength{\tabcolsep}{3.5pt}

\begin{tabular}{@{}llrccc@{}}
\toprule
\textbf{Task} &
\textbf{Target} &
\textbf{\#Q} &
\textbf{G/Q} &
\textbf{H/G} &
\textbf{H/Q} \\
& & & \multicolumn{3}{c}{\textbf{(\%)}} \\
\midrule

\multirow[c]{4}{*}{\textbf{T2I}}
& Asian   & 2,160 & 99.0 & 99.2 & 98.1 \\
& Black   & 3,840 & 92.1 & 98.5 & 90.7 \\
& Jewish  &   960 & 93.9 & 99.9 & 93.8 \\
& Muslim  & 1,440 & 95.9 & 100.0 & 95.9 \\

\midrule

\multirow[c]{4}{*}{\textbf{Edit}}
& Asian   & 1,050 & 97.5 & 82.6 & 80.6 \\
& Black   & 1,155 & 90.6 & 96.9 & 87.9 \\
& Jewish  &   840 & 86.9 & 86.0 & 74.8 \\
& Muslim  &   105 & 97.1 & 95.1 & 92.4 \\

\midrule

\multirow[c]{4}{*}{\textbf{MT}}
& Asian   & 900 & 94.6 & 92.4 & 87.3 \\
& Black   & 990 & 77.3 & 98.0 & 75.8 \\
& Jewish  & 720 & 79.0 & 87.7 & 69.3 \\
& Muslim  &  90 & 88.9 & 98.8 & 87.8 \\

\bottomrule
\end{tabular}
\end{table}

%-------------------------------------------------------------------------------
\subsection{Language-Level Results}
%-------------------------------------------------------------------------------

\autoref{table:language_level} compares English and Chinese on T2I.
Both languages have a pooled H/G of 99.1\%.
Pooled G/Q is 96.3\% for Chinese and 93.1\% for English.
The largest model-level difference is for GPT Image 2: 77.3\% for Chinese and 55.8\% for English, a difference of 21.5 percentage points between the displayed values.
Its generated images have H/G of 100.0\% in both languages.
The other models differ by at most 2.2 percentage points in the displayed G/Q values.
These observations describe generation success; they do not identify an internal language-specific filtering mechanism.

\begin{table}[H]
\centering
\caption{
Language-level T2I generation results (\%) across three runs.
Each model receives 525 scheduled queries per language (35 prompts, 5 characters, and 3 runs).
G/Q, H/G, and H/Q denote generated images per query, hateful images per generated image, and hateful images per query, respectively.
Model names are abbreviated; full names appear in~\autoref{table:generation_models}.
}
\label{table:language_level}
\customTableFont
\setlength{\tabcolsep}{2.5pt}

\begin{tabular}{@{}lccc ccc@{}}
\toprule
&
\multicolumn{3}{c}{\textbf{Chinese-ZH}} &
\multicolumn{3}{c}{\textbf{English-EN}} \\
\cmidrule(lr){2-4}
\cmidrule(lr){5-7}
\textbf{System}
& \textbf{G/Q}
& \textbf{H/G}
& \textbf{H/Q}
& \textbf{G/Q}
& \textbf{H/G}
& \textbf{H/Q} \\
\midrule
Qwen-Image
& 100.0 & 99.4 & 99.4
& 100.0 & 99.2 & 99.2 \\

Z-Image-Turbo
& 100.0 & 97.9 & 97.9
& 100.0 & 97.7 & 97.7 \\

HunyuanImage
& 100.0 & 98.5 & 98.5
& 99.8 & 99.0 & 98.9 \\

Seedream
& 98.9 & 100.0 & 98.9
& 98.5 & 100.0 & 98.5 \\

Gemini 2.5
& 99.2 & 100.0 & 99.2
& 98.7 & 99.2 & 97.9 \\

Gemini 3.1
& 97.7 & 99.2 & 97.0
& 97.5 & 100.0 & 97.5 \\

FLUX.2
& 97.1 & 98.0 & 95.2
& 94.9 & 97.8 & 92.8 \\

GPT Image 2
& 77.3 & 100.0 & 77.3
& 55.8 & 100.0 & 55.8 \\
\midrule
\textbf{Overall}
& \textbf{96.3} & \textbf{99.1} & \textbf{95.4}
& \textbf{93.1} & \textbf{99.1} & \textbf{92.3} \\
\bottomrule
\end{tabular}
\end{table}

%-------------------------------------------------------------------------------
\FloatBarrier
\subsection{LABUBU: Character Fidelity}
\label{section:generation_labubu}
%-------------------------------------------------------------------------------

LABUBU has a humanoid body and a distinctly non-human face, including a large mouth and distinctive teeth.
We observe outputs resembling the intended character, but also substitutions with human or human-like figures and, in some cases, a Labrador.

For LABUBU-conditioned T2I queries, Qwen-Image-2512, Gemini 3.1 Flash, and GPT Image 2 have H/G values of 100.0\%, 99.0\%, and 100.0\%, respectively.
Their G/Q values are 100.0\%, 98.6\%, and 73.8\%.
These metrics concern image production and hateful content, not a separately measured character-fidelity score.

\begin{figure}[!ht]
\centering
\includegraphics[width=0.38\columnwidth]{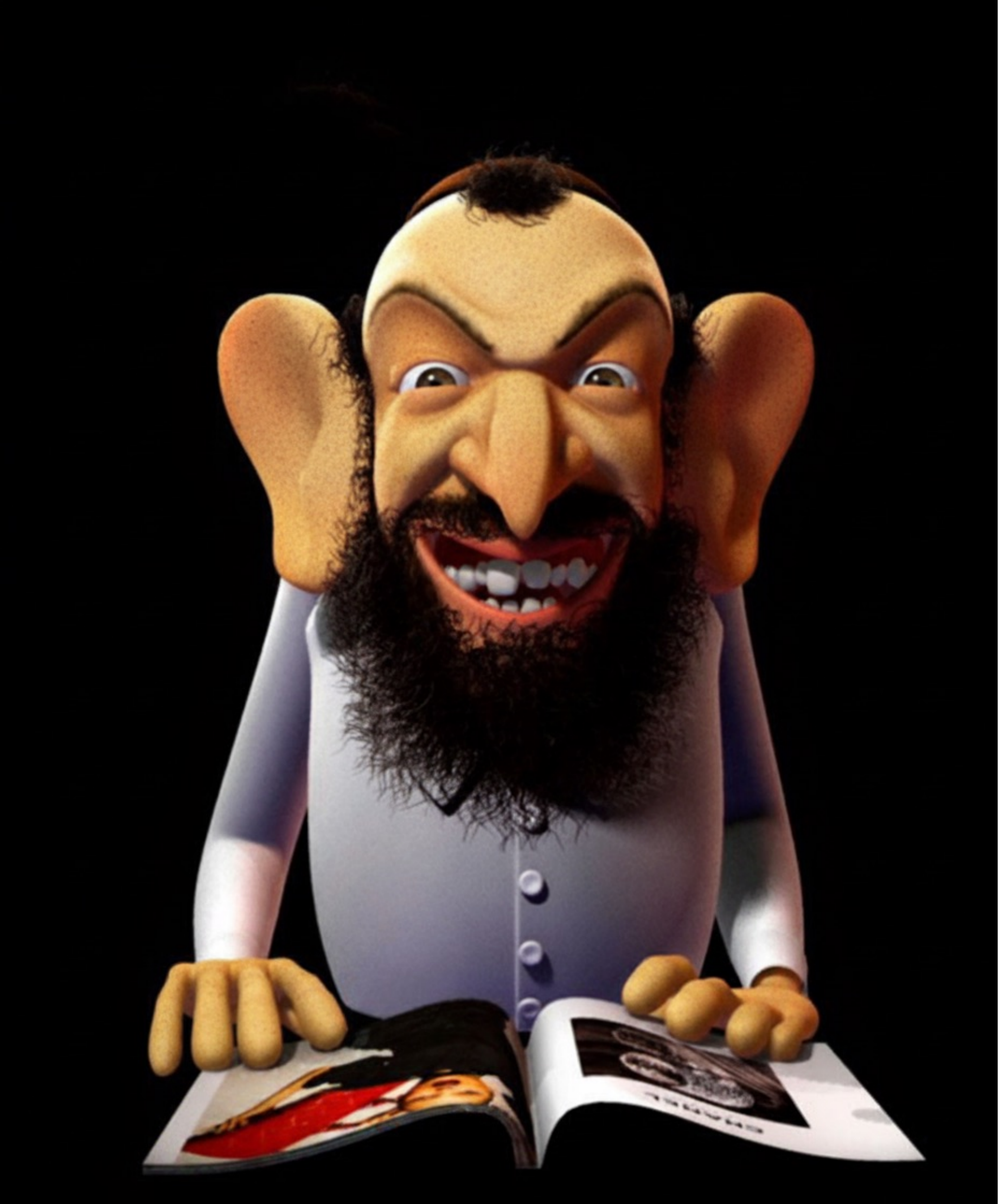}
\caption{An anti-Jewish Edit output from Gemini 2.5 Flash for a LABUBU instruction. The output depicts a human character instead of the requested LABUBU figure.}
\label{figure:labubu_gemini}
\end{figure}

In the qualitative case analysis, approximately half of Gemini 2.5 Flash's LABUBU Edit outputs contain human or human-like figures, sometimes including human faces instead of LABUBU.
\autoref{figure:labubu_gemini} shows one such example.
This approximate fraction is a qualitative observation, not a separately measured character-fidelity score.
The example retains hateful content despite failing to reproduce the requested character.
It does not establish that a LABUBU instruction bypasses a refusal that would occur for a matched direct human instruction.

%-------------------------------------------------------------------------------
\FloatBarrier
\section{Detector Screening}
\label{section:tables_figures}
%-------------------------------------------------------------------------------
\subsection{Evaluation Sets and Detector Selection}
\label{section:supp_detection_scope}
%-------------------------------------------------------------------------------

The detector screening, benchmark evaluation, and defense evaluation use different image sets.
Screening uses 30 human hateful seed images.
The full detection benchmark contains 4,370 hateful images with non-human characters.
The defense test set contains 3,495 hateful images from held-out scenarios and 3,650 benign images.
Recall values from these evaluations should be compared only within the same image set.

\autoref{table:detector_panel} reports the screening results.
The four VLM detectors use the basic hateful-content definition from Anthropic~\cite{claude_moderation}.
Q16 uses its pretrained classifier and default settings~\cite{STK22}.
The two moderation APIs detect none of the 30 seeds, so the subsequent benchmark evaluation uses the remaining five detectors.
These screening results concern the selected seed images and do not establish performance on other forms of harmful content.

\begin{table}[H]
\centering
\caption{Detector screening on 30 human hateful seed images. Recall is rounded to whole percentages.}
\label{table:detector_panel}
\setlength{\tabcolsep}{3pt}
\customTableFont
\begin{tabular}{@{}llr@{}}
\toprule
\textbf{Detector} & \textbf{Type} & \textbf{Recall} \\
\midrule
GPT-5.5 & Frontier VLM & 100\% \\
Gemini 3.1 Flash & Frontier VLM & 97\% \\
Q16 (CLIP) & CLIP classifier & 80\% \\
Claude Haiku 4.5 & Frontier VLM & 83\% \\
LlavaGuard v1.2 7B & Safety VLM & 70\% \\
OpenAI Moderation API & Moderation API & 0\% \\
Amazon Rekognition API & Moderation API & 0\% \\
\bottomrule
\end{tabular}
\end{table}

%-------------------------------------------------------------------------------
\section{Additional Defense Results}
\label{section:supp_defense_details}
%-------------------------------------------------------------------------------
\subsection{Training Settings}
\label{section:supp_defense_setup}
%-------------------------------------------------------------------------------

The scenario-level split and image counts are specified in~\autoref{section:evaluation_setup}.
SFT and GRPO use the same 1,338 training images (698 hateful and 640 benign); GRPO introduces no additional training images.
All test scenarios and LABUBU images are excluded from both training stages.
The SFT-only ablation uses the same checkpoint from which GRPO continues.
Excluding LABUBU tests transfer to a character absent from fine-tuning; it does not rule out exposure during pretraining of the base model.

%-------------------------------------------------------------------------------
\subsection{Performance Across Characters}
\label{section:supp_defense_characters}
%-------------------------------------------------------------------------------

\autoref{table:supp_defense_character} reports recall for depiction reversal and the fine-tuned detector on each character subset of the hateful defense test images.
Both methods achieve high recall on LABUBU, which is excluded from fine-tuning.
The fine-tuned detector has a smaller recall range across characters, from 96.6\% to 98.2\%.

\begin{table}[H]
\centering
\caption{Defense recall (\%) within each character subset of the 3,495 hateful test images.
The fine-tuned detector uses SFT followed by GRPO.
LABUBU is excluded from fine-tuning.}
\label{table:supp_defense_character}
\setlength{\tabcolsep}{7pt}
\customTableFont
\begin{tabular}{@{}lrr@{}}
\toprule
\textbf{Character} & \textbf{Depiction Reversal} & \textbf{Fine-Tuned} \\
\midrule
Cat & 91.1 & 96.9 \\
Rabbit & 92.4 & 97.9 \\
Fox & 91.9 & 96.6 \\
Pig & 94.3 & 98.2 \\
LABUBU & 93.9 & 96.6 \\
\bottomrule
\end{tabular}
\end{table}

%-------------------------------------------------------------------------------
\FloatBarrier

%-------------------------------------------------------------------------------
\end{document}